\documentclass{ieeeaccess}
\usepackage{cite}
\usepackage{amsmath,amssymb,amsfonts}
\usepackage{algorithmic}
\usepackage{graphicx}
\usepackage{textcomp}
\def\BibTeX{{\rm B\kern-.05em{\sc i\kern-.025em b}\kern-.08em
    T\kern-.1667em\lower.7ex\hbox{E}\kern-.125emX}}

 \usepackage{multirow}
 \usepackage[english]{babel}

\usepackage{amsthm}
 \newtheorem{proposition}{Proposition}  
\DeclareUnicodeCharacter{2212}{\ensuremath{-}}
\usepackage{hyperref}
\usepackage{bm} 
\hypersetup{
    colorlinks=true,
    linkcolor=blue,
    filecolor=magenta,      
    urlcolor=cyan,
    pdftitle={Overleaf Example},
    pdfpagemode=FullScreen,
    }
\begin{document}
\history{Date of publication xxxx 00, 0000, date of current version xxxx 00, 0000.}
\doi{10.1109/ACCESS.2017.DOI}

\title{MQ-GNN: A Multi-Queue Pipelined Architecture for Scalable and Efficient GNN Training}
\author{\uppercase{Irfan Ullah}\authorrefmark{1}, and 
\uppercase{Young-Koo Lee\authorrefmark{2}\IEEEmembership{Member, IEEE}}
}
\address[1]{Department of Computer Science and Engineering, Kyung Hee University (Global Campus), Republic of Korea (e-mail: irfanr@khu.ac.kr)}
\address[2]{College of Software, Kyung Hee University (Global Campus), Republic of Korea (e-mail: yklee@khu.ac.kr)}

\markboth
{Irfan \headeretal: MQ-GNN}
{Irfan \headeretal: MQ-GNN}

\corresp{Corresponding author: Young-Koo Lee (e-mail: yklee@khu.ac.kr).}

\begin{abstract}
Graph Neural Networks (GNNs) are powerful tools for learning graph-structured data, but their scalability is hindered by inefficient mini-batch generation, data transfer bottlenecks, and costly inter-GPU synchronization. Existing training frameworks fail to overlap these stages, leading to suboptimal resource utilization. This paper proposes MQ-GNN, a multi-queue pipelined framework that maximizes training efficiency by interleaving GNN training stages and optimizing resource utilization. MQ-GNN introduces Ready-to-Update Asynchronous Consistent Model (RaCoM), which enables asynchronous gradient sharing and model updates while ensuring global consistency through adaptive periodic synchronization. Additionally, it employs global neighbor sampling with caching to reduce data transfer overhead and an adaptive queue-sizing strategy to balance computation and memory efficiency. Experiments on four large-scale datasets and ten baseline models demonstrate that MQ-GNN achieves up to \boldmath $\bm{4.6\,\times}$ faster training time and \textbf{$30\%$} improved GPU utilization while maintaining competitive accuracy. These results establish MQ-GNN as a scalable and efficient solution for multi-GPU GNN training. The code is available at \href{https://github.com/sahibzada-irfanullah/MQ-GNN}{MQ-GNN}. 
\end{abstract}

\begin{keywords}
Graph neural network, Multi-GPU, pipeline, optimization, Mixed CPU-GPU  training, GNN Training, Staleness, Inter-GPU communication
\end{keywords}

\titlepgskip=-15pt

\maketitle

\section{Introduction}
\label{sec:introduction}
\PARstart{G}{raphs} naturally represent complex relationships in many real-world applications, such as social networks, biological systems, and knowledge graphs \cite{chami2022machine}. By learning low-dimensional embeddings of graph nodes, GNNs efficiently represent graph information, enabling a wide range of downstream applications, such as node classification \cite{hamilton2017inductive,velivckovic2017graph} and link prediction \cite{zou2019layer, chen2018fastgcn, chami2022machine}. A diverse array of models \cite{hamilton2017inductive,velivckovic2017graph} has demonstrated state-of-the-art performance across a variety of domains, including protein structures \cite{strokach2020fast}, social networks \cite{fan2019graph}, and knowledge graphs \cite{park2019estimating}.

Despite their success, training GNNs on large-scale graphs in a distributed GPU environment presents significant challenges due to memory constraints, computational costs, and the interdependencies inherent in graph structures. Real-world graphs often contain millions of nodes and edges, along with high-dimensional node and edge features \cite{gandhi2021p3, thorpe2021dorylus}. These factors result in substantial storage and computational overheads. Specific datasets may require numerous GBs of storage \cite{ilyas2022saga}, making it inefficient or even infeasible to consider all neighbors within $L$ hops for each training node as a single batch when training a GNN model with $L$ layers \cite{lin2020pagraph}. Furthermore, most graphs exhibit highly skewed degree distributions \cite{chen2019powerlyra}, where a small number of well-connected nodes aggregate features from a substantial portion of the graph within just a few hops. These challenges necessitate mini-batch training, which reduces computational and memory overhead by processing smaller samples iteratively \cite{zhang2020deep}.

For mini-batch training, sampling-based approaches such as node-wise sampling \cite{hamilton2017inductive, chen2018stochastic} and layer-wise sampling \cite{chen2018fastgcn, huang2018adaptive, zou2019layer} have been developed. Node-wise and layer-wise sampling are among the most widely used and practical approaches for training GNNs, each addressing unique scalability and computational efficiency challenges. Node-wise sampling selects a fixed number of neighbors for each node. However, it introduces additional computational challenges, particularly when the mini-batch size increases. For instance, computing embeddings in $L$-layer GNNs requires aggregating information from $L$-hop neighbors, leading to the well-known "neighbor explosion" problem \cite{zhen2020graphsaint, chen2018stochastic, huang2018adaptive}. To address this issue, layer-wise sampling was developed. It selects subsets of nodes at each layer for sample generation. However, both node-wise and layer-wise approaches require frequent data transfers between CPU and GPU memory, reducing training efficiency. To reduce frequent and redundant data transfers, global neighbor sampling (GNS) \cite{dong2021global} periodically samples a small number of nodes for all mini-batches and caches their features in GPU memory. GNS prioritizes neighbors in the cache, reducing distinct nodes per batch, increasing overlap among them, and minimizing CPU-GPU data transfers to speed up training. However, sample generation, data transfer, and computation do not overlap efficiently. The lack of efficient queues and pipelines and poor interleaving of GNN training tasks such as mini-batch preparation, data transfer, computation, and updates leads to significant latency and underutilized computational resources. This issue is particularly severe for large-scale graphs, where sampling and data movement dominate the runtime, causing GPUs to starve for data and stalling the training process.

Recently, multiple GPUs have been extensively used to accelerate GNN training \cite{huang2021understanding, jia2020improving}. In a typical single-machine multi-GPU setup, large-scale graphs (both topology and feature data) exceed limited GPU memory capacity (e.g., 12 GB on an NVIDIA 3060), so most existing GNN systems store topological and feature data in host memory. CPUs repeatedly sample input graphs, extract features for sampled nodes, and transfer them to GPUs for training. Unfortunately, this process incurs high data transfer costs, dominating training and significantly underutilizing GPUs. This occurs because GNN models often use deep neural networks with minimal computational complexity relative to large data volumes. As a result, GNN computation on a GPU is substantially faster than data loading. For instance, \cite{lin2020pagraph} reported that data loading consumed 74\% of training time. Existing GNN training frameworks, such as Deep Graph Library (DGL) \cite{wang2019dgl} and PyTorch Geometric (PyG) \cite{min2021pytorch}, partially interleave processes such as mini-batch generation, data transfer, and computation. However, these frameworks still exhibit significant inefficiencies, as the interleaving cannot overlap these tasks fully. GPUs frequently become idle while waiting for the CPU to complete sampling and data transfer, resulting in poor computational resource utilization (up to 50\%). These bottlenecks are aggravated in larger graphs, leading to redundant accesses to nodes, increased communication costs, and increased training time. In multi-GPU systems, the demand for data samples increases linearly, increasing communication and synchronization overheads and further limiting scalability. The system must generate and distribute more mini-batches to keep all GPUs active as the GPU count increases. This increases the system's burden, including generating samples, fetching CPU features, and moving them to GPUs. The communication volume also grows with more GPUs, leading to more significant overheads and delays. Each GPU needs to exchange gradients or intermediate results with other GPUs to update the models. Coordinating model updates and gradients across GPUs requires frequent synchronization. As the number of GPUs increases, the cost of this synchronization rises, resulting in bottlenecks where GPUs sit idle while others catch up.

To overcome these limitations, we propose MQ-GNN, a multi-queue pipelined architecture. This versatile and scalable pipeline architecture optimizes resource utilization by interleaving mini-batch generation, data transfer, GNN computation, gradient sharing, and model updating. MQ-GNN achieves this by employing multi-queues to manage mini-batches in CPU and GPU memory, along with gradients in GPU memory. By leveraging mini-batch queues in the main memory (one per GPU), MQ-GNN ensures concurrent data transfer and computation, minimizing GPU idle times. It is designed to work efficiently with any node-wise or layer-wise sampling approach, providing robust training performance regardless of the sampling method. Additionally, MQ-GNN employs global neighbor sampling to periodically cache shared nodes across mini-batches, reducing redundant accesses and CPU-GPU data transfers to accelerate training.

Furthermore, MQ-GNN introduces the Ready-to-Update Asynchronous Consistent Model (RaCoM), which uses gradient queues to enable asynchronous gradient sharing between GPUs. This allows gradients to be accumulated while models are updated concurrently with the next mini-batch computation. MQ-GNN employs periodic synchronization with intervals adapted to dataset sparsity or density to ensure model consistency and mitigate divergence. This approach balances asynchronous gradient sharing and updates, which decreases staleness and improves training efficiency.

Determining the optimal queue size in a multi-queue pipeline is crucial for balancing computational efficiency and memory usage. A large queue inflates memory usage and may cause overflows, whereas a small queue increases latency and underutilizes resources. This trade-off becomes critical when training on large datasets with diverse memory and computational demands. We propose a method to determine the optimal queue size by (1) integrating sampling and data transfer time with GPU compute time and (2) capping it at peak memory usage. This ensures feasibility for memory-intensive datasets while maximizing GPU utilization within memory constraints.

Finally, we investigated both node-wise and layer-wise sampling approaches to show MQ-GNN’s effectiveness and adaptability to different sampling approaches in GNN training. Furthermore, we conducted a comprehensive multi-GPU analysis of both sampling approaches, examining how GPU count, dataset variability, and staleness affect performance. To the best of our knowledge, this is the first work that assesses layer-wise sampling in multi-GPU training within a pipelined, multi-queued framework, demonstrating MQ-GNN’s scalability and efficiency across diverse configurations.

In summary, the key contributions of this paper are:

\begin{enumerate}

\item Proposing MQ-GNN, a multi-queue pipelined architecture, optimizes CPU, memory, and GPU utilization by interleaving mini-batch generation, data transfer, computation, and gradient sharing. This approach reduces data movement latency and maximizes GPU efficiency during GNN training.
\begin{itemize} 

    \item To improve GPU utilization, a method to determine the optimal queue size based on available memory and processing time is introduced.
    \item  Employing global neighborhood sampling to cache frequently accessed nodes, reducing redundant accesses and minimizing CPU-GPU data transfers.

    \end{itemize}

\item Proposing RaCoM, a framework for asynchronous gradient sharing that uses gradient queues to enable independent local model updates on GPUs, reducing synchronization delays.

    \begin{itemize} \item Introducing a periodic synchronization mechanism to improve model consistency by balancing communication overhead and gradient staleness based on graph sparsity or density, enhancing scalability and robustness.
    \end{itemize}

 \item Conducting a comprehensive empirical analysis of node-wise and layer-wise sampling in multi-GPU settings, providing the first evaluation of layer-wise sampling performance and staleness in a multi-GPU setup, offering insights into scalability and efficiency.

\end{enumerate}

\section{Related Work}
The growing scale of graph datasets has led to the development of efficient and scalable training paradigms for Graph Neural Networks (GNNs). This section is devoted to discussing and reviewing the existing literature.

\subsection{Scalable GNN Training}
Several systems have been designed to facilitate GPU-based GNN training \cite{dong2021global,kaler2022accelerating,wu2020comprehensive}. Frameworks such as Deep Graph Library (DGL) \cite{wang2019dgl} and PyTorch Geometric (PyG) \cite{min2021pytorch} utilize CPU memory for graph storage and enable distributed mini-batch training across multiple GPUs. However, these frameworks often face inefficiencies, including GPU underutilization, sub-linear speedups in multi-GPU setups \cite{zheng2022distributed}, and high data transfer overhead. To overcome these challenges, PaGraph \cite{lin2020pagraph} minimizes data transfer overhead with computation-aware caching, while DSP \cite{van2018dspefficient} improves multi-GPU training through dynamic graph partitioning and synchronized mini-batches. MSPipe \cite{sheng2024mspipe} optimizes temporal GNNs using pipelined execution and memory-efficient caching to mitigate staleness.

MQ-GNN is different in that it incorporates mini-batch generation, data transfer, computation, and gradient sharing into a pipelined architecture with multi-queues.

\subsection{Graph Sampling Methods}

Graph sampling methods are crucial for addressing challenges such as neighbor explosion and memory constraints in GNN training. Node-wise sampling (NS) approaches, such as fixed neighbor sampling in GraphSAGE \cite{hamilton2017inductive}, select neighbors independently for each node, reducing computation but introducing redundancy in embedding calculations \cite{huang2018adaptive}. Other studies propose layer-wise (LS) sampling approaches, such as FastGCN \cite{chen2018fastgcn}. FastGCN samples a specified number of nodes per layer based on probabilities derived from each node's degree. GNN generalization and training performance is affected because sampled nodes in consecutive layers may not be connected, as their sampling probabilities are computed independently. To ensure quick and smooth convergence, the sampling probability should ideally be computed to lower the estimation variance in FastGCN \cite{chen2018fastgcn}. As a result, the adjacency matrix can become highly sparse or contain all-zero rows, leading to disconnected nodes and an imprecise computation graph, ultimately degrading FastGCN's training and generalization performance. Huang et al. \cite{huang2018adaptive} presented an adaptive and trainable sampling approach that conducts LS conditioned on the former layer to capture the inter-layer correlation and lower the estimation variance. It achieved higher accuracy than FastGCN at the cost of using a much more complicated sampling approach. Zou et al. \cite{ladies2019} proposed LAyer-Dependent Importance Sampling (LADIES), an efficient sampling algorithm that builds on the strengths and weaknesses of previous approaches. LADIES further reduces training variance and aims to mitigate sparse connections in FastGCN. To improve scalability in sampling-based GCN training, Chen et al. \cite{chen2023calibrate} focuses on history-oblivious LS methods (e.g., FastGCN and LADIES), which construct sampling probabilities without relying on historical data. They revisit this approach from a matrix approximation perspective and address two key issues: suboptimal sampling probabilities and estimation biases caused by sampling without replacement in existing LS methods.

MQ-GNN integrates these advanced sampling techniques with global neighbor sampling, periodically caching frequently accessed nodes across mini-batches. This reduces redundant data access, alleviates memory bottlenecks, and accelerates training while maintaining high accuracy. It supports NS and LS strategies, ensuring efficient training regardless of the sampling method.

\subsection{Pipeline-Based Architectures}
Pipelined architectures effectively improve GPU utilization by overlapping data movement and computation. Marius \cite{mohoney2021marius} and MSPipe \cite{sheng2024mspipe} demonstrate the benefits of pipelining for graph embeddings and temporal GNNs, respectively. However, these systems primarily focus on small graphs or specialized application domains, limiting their scalability. DSP \cite{van2018dspefficient} optimizes pipelined execution for multi-GPU setups with synchronized training but depends heavily on graph partitioning for performance. Marius \cite{mohoney2021marius} employs queues for mini-batches but lacks support for GNN training, as it is designed for non-GNN graph embedding.

Communication overhead is a significant bottleneck in multi-GPU GNN training. GNNPipe \cite{chen2023gnnpipe} and PipeGCN \cite{wan2022pipegcn} tackle this issue in full-batch training scenarios. GNNPipe addresses this by introducing layer-level model parallelism, partitioning GNN layers across GPUs to reduce communication volume proportionally to the number of layers. Each GPU processes the entire graph for its assigned layers, utilizing historical embeddings and specialized training techniques to ensure convergence. While GNNPipe reduces communication and training overhead, its reliance on full-graph processing limits scalability for large graphs. Its reliance on layer partitioning also reduces flexibility for integrating sampling-based methods. PipeGCN reduces inter-partition communication overhead by overlapping communication and computation. In contrast, MQ-GNN eliminates full-graph dependencies through a sampling-based framework. Its pipelined architecture minimizes communication overhead without full-graph processing. MQ-GNN adapts to diverse datasets and computational setups by supporting node- and layer-wise sampling.

MQ-GNN introduces a multi-queue-based pipeline for large-scale GNN training without partitioning the graph. It interleaves mini-batch generation, data transfer, and gradient sharing, improving efficiency. Dedicated queues for mini-batches and gradients mitigate staleness, while periodic synchronization ensures consistency. This design enables efficient GNN training on both dense and sparse graphs, addressing bottlenecks in data movement, computation, and inter-GPU communication.

\section{Preliminaries and Notations} \label{sec:prelimAndNotat}
This section introduces the necessary notation and background for various sampling methods in GNN training.

Let \(\mathbf{G} = (V, E)\) be a graph. The node set is defined as \( V = \{ v_i \mid i \in [n] \} \), where \([n] = \{0, 1, 2, \dots, n-1\}\) represents the index set of nodes. The edge set is given by \( E = \{ (v_i, v_j) \mid i, j \in [n] \} \). An edge \( (v_i, v_j) \in E \) represents a connection between nodes \( v_i \) and \( v_j \).  Each node \( v_i \) has a feature vector \( f_i \in \mathbb{R}^d \), where \( d \) is the feature dimension. All node feature sets are denoted as \( F = \{ f_i \mid v_i \in V \} \). Each node \( v_i \) is associated with a label \( y_i \in \mathcal{Y} \), where \( \mathcal{Y} \) is the set of possible labels. The label set for all nodes is $\mathcal{Z} = \{ y_i \mid v_i \in V \}$ where \( y_i \) represents the label of node \( v_i \).

Many sampling methods can be employed during GNN training. Sampling is very important for creating mini-batches, which enhances training throughput and accelerates model convergence.  These sampling methods, based on the GNN layer and node neighborhood, are as follows:

\subsection{Node-wise Sampling (NS)}\label{subsec:ns}
NS selects a subset of neighbors for each node to construct smaller, computationally manageable subgraphs for training. Its primary goal is to reduce memory and computational costs while preserving the graph's local neighborhood structure.

\textbf{GCN} For a given graph \textbf{G}, the $l$-th convolution layer in GCN can be defined as:

\begin{equation}
\label{eq:gcn_eq}
Z^l = PH^{l-1}W^{l-1}, H^{l-1} = \sigma(Z^{l-1}), 
\end{equation}

where $L$ is the total number of layers, $l \in \{1, \dots, L\}$ is the layer index, $\sigma$ is an activation function, $H^{l-1}$ represents the embedding at layer $l-1$,  $Z^{l-1}$ is the intermediate embedding at layer $l-1$, and $W^{l-1}$ denotes the weight matrix. $P$ is the normalized Laplacian matrix, defined as:

\begin{equation}
\label{eq:norm_lap_mat}
P = \widehat{D}^{-1/2} \widehat{A} \widehat{D}^{-1/2}, \widehat{D}_{i, j} = \sum_{j} \widehat{A}_{i, j},
\end{equation}

where $\widehat{D}$ is a diagonal matrix, and $\widehat{A}$ represents a normalization of adjacency matrix $A$, i.e., $\widehat{A}=A + I$.

\textbf{GraphSAGE} GraphSAGE \cite{hamilton2017inductive} uses neighbor sampling to control the receptive field size in GNNs. For each node at the \( l \)-th layer, a fixed number of its neighbors, denoted as \( s_{node} \), are randomly and uniformly sampled. This approach formulates an unbiased estimator, \( \widehat{P}^{l-1}H^{l-1} \), to approximate \( PH^{l-1} \) in the graph convolution layer and optimize computation and memory efficiency \cite{chen2023calibrate}.

\begin{equation}
\label{eq:sage_eq}
\widehat{P}_{i, j}^{l-1} =
\begin{cases}
\frac{\left| \mathcal{N}(v_i) \right|}{s_{node}} P_{i, j}, & \text{if } v_j \in \mathcal{\widehat{N}}^{l-1}(v_i)   \\ 
0, & \text{otherwise},
\end{cases}
\end{equation}

where $\mathcal{N}(v_i)$ and $\mathcal{\widehat{N}}^{l-1}(v_i)$ denote the full and sampled neighbor sets of node $v_i$ at the $(l-1)-th$ layer.

\subsection{Layer-wise sampling (LS)} \label{subsec:ls}
Instead of sampling a fixed set of neighbors for each node, LS selects a globally defined set of nodes at each layer. LS offers broader receptive field coverage and reduces variation in node representations compared to NS. By adopting a layer-wide perspective, LS balances computational cost and information coverage, making it well-suited for large graphs and tasks requiring robust layer-level feature aggregation. However, LS may be less effective for tasks that depend on preserving fine-grained local neighborhood structures.

\textbf{FastGCN}\label{sec:fastgcn} Instead of sampling neighbors for each node, FastGCN \cite{fey2019fast} selects nodes at the layer level, modeling each graph layer as an embedding function over nodes, which are treated as random variables under a probability measure \( P \). Importance sampling probabilities determine how nodes are prioritized for selection during training, reducing the variance in training outcomes. Each node \( i \) is assigned a probability \( p_i \propto \| RP_i \|^2\) for \( i \in [n] \), where $R$ is the row selection matrix, assuming independence. For a given layer \( l \), \( r_l \) i.i.d. samples \( \{ u_1^l, \dots, u_{r_l}^l \} \sim P \) are drawn to approximate the embeddings for all nodes in the layer, using the following formulation \cite{chen2018fastgcn,fey2019fast}:

\begin{equation}
\begin{aligned}
\label{eq:fastgcn}
\widehat{z}_{r_{l+1}}^{l+1}\left ( v \right ) =  \frac{1}{r_l}  \sum_{i=1}^{r_l} \widehat{A}\left ( v, u_{i}^{l} \right  ) z_{r_l}^{l} \left ( u_{i}^{l}\right ) W^l, \\ 
h_{r_l+1}^{l+1} \left ( v \right ) = \sigma \left (\widehat{z}_{r_{l+1}}^{l+1} \left ( v \right ) \right), \quad l = 0, 1,..., L-1,
\end{aligned}
\end{equation}

where \( h_{r_l+1}^{l+1} \) represents the embedding at layer \( l+1 \), \(z_{r_0}^{0} = z^0\) denotes the node features ($f$), \( \widehat{A}\left ( v, u_{i}^{l} \right  ) \) corresponds to the \((v, u)\) element of \( \widehat{A} \). The loss function with function \( g \) can be estimated as:

\begin{equation}
\text{loss}_{r_0, r_2, \dots, r_L} = \frac{1}{r_L}  \sum_{i=1}^{r_L} g \left ( z_{r_L}^{L}(u_{i}^{r_L}) \right  ).
\end{equation}

\textbf{LADIES} \label{sec:ladies} Independently performing layer-wise sampling at different layers can be inefficient, as the resulting bipartite graph may be sparse or even contain all-zero rows. LADIES \cite{ladies2019} modifies independent layer-wise sampling to construct the computation graph in FastGCN training to address this issue. In their approach, at each layer, nodes are sampled from the union of the neighbors of the previously sampled nodes, as given by

\begin{equation}
\label{eq:ladies}
\mathcal{V}^{l-1} = \bigcup_{v_i \in \mathcal{S}_l} \mathcal{N}(v_i), 
\end{equation}

where \( \mathcal{V}^{l-1} \) represents the sampled nodes at layer \( l-1 \), \( \mathcal{S}_l \) is the set of nodes sampled at the \( l \)-th layer, and \( \mathcal{N}(v_i) \) denotes the set of neighbors of node \( v_i \). Therefore, during the sampling process, probabilities are assigned only to nodes in $\mathcal{V}^{l-1}$, given by $\left\{p_{i}^{l-1} \right \}_{v_{i} \in \mathcal{V}^{l-1}}$. Specifically, the selection probabilities for nodes are defined as:

\begin{equation}
\label{eq:ladiesPro}
p_{i}^{l-1} = \frac{\left \| R^{l}P_{*,i} \right \|_2^2}{\left \| R^{l}P \right \|_F^2},
\end{equation}

where $R^l$ is the row selection matrix. Assuming that the sets of sampled nodes are determined at layers \( l \) and \( l-2 \), and that each node $v_i$ is assigned probabilities $p_{1}^{l-1}...p_{|V|}^{l-1}$, the diagonal matrix $S_{s,s}^{l-1}$ is defined as:

\begin{equation}
  S_{s,s}^{l-1} =
  \begin{cases}
    \frac{1}{s_{l-1} p_{i_k^{l-1}}^{l-1}}, & s = i_k^{l-1}, \\ 
    0, & \text{otherwise}.
  \end{cases}
\end{equation}

Here $s_{l-1}$ is the number of sampled nodes at layer $l-1$. Due to the absence of information on the intermediate embedding or activation matrix during the characterization of the samples at the $1-th$ layer, an essential sampling scheme is employed, relying solely on the row selection matrices $R^l$ and a normalized Laplacian matrix $L$. The row selection matrix $R^l$ can be defined as:

\begin{equation}
R^l_{c,r} = \begin{cases}
1, & (c, r) = (c, i_{c}^{l}),\\ 
 0 ,& \text{ otherwise.} 
\end{cases}
\end{equation}

where ${i_c^l }$  represents the sets of nodes at $S_l$. Meanwhile, $L$ is defined as, $L=\widehat{D}^{-1/2}\widehat{A}\widehat{D}^{-1/2}$.  Finally, the sampled training mini-batch is defined in terms of the Laplacian adjacency matrix and corresponding embeddings (or features at layer 0) as \( \frac{1}{p_{k}^{l-1}}\widehat{L}_{*,i} \) and \( \widehat{Z}_{k,*}^{l-1} \), where  $\widehat{L} \leftarrow D_{\widehat{p}^l}^{l-1}\widehat{p}^l$ is the normalized Laplacian matrix, avoiding vanishing and exploding gradients.

\textbf{Calibrate and Debias LS} The efficiency of LS heavily depends on the importance of the sampling procedure, which estimates node aggregations using significantly fewer nodes while preserving accuracy. The selection of sampling probabilities is crucial, as it directly impacts the accuracy of GCNs \cite{chen2023calibrate}. To improve the scalability of sampling-based GCN training, Chen et al. \cite{chen2023calibrate} investigated history-oblivious layer-wise sampling techniques, such as FastGCN and LADIES, which construct sampling probabilities without relying on historical data. By analyzing these methods from a matrix approximation perspective, they identified and addressed two key issues: suboptimal sampling probabilities and estimation biases caused by sampling without replacement.

The weak or negative correlation between \( \| H W_i \| \) (where \( W_i \) is the weight matrix) and \( \| P^i \| \) highlights the limitations of the proportionality assumption in FastGCN and LADIES sampling probabilities. The authors recognize the constrained prior knowledge of \( H W \) in the history-oblivious setting to address this limitation. Guided by the Principle of Maximum Entropy, they assume a uniform distribution for \( \| HW_i \| \). Based on this assumption, they derive the following sampling probabilities, referred to as 'flat' sampling \cite{chen2023calibrate}:

\begin{equation}
\label{eq:calibProFlat}
 p_{i} \propto \left \| HW_i \right \|, \quad \forall i \in [n].
\end{equation}

The sampling probabilities are adjusted to balance variance and accuracy by reformulating the target matrix product as \( R P I (H W) \), where \( R P I \) is approximated. Assuming a uniform distribution for the norms of rows in \( HW \) improves both the stability of the matrix approximation and the prediction accuracy of GCNs, providing a more robust and reliable sampling framework. The advanced variants of LADIES and FastGCN that incorporate this approach are referred to as LADIES+flat (or LADIES+f) and FastGCN+flat (or FastGCN+f), respectively \cite{chen2023calibrate}.

Although the probabilities in \eqref{eq:calibProFlat} improve performance, they assume that neighbor nodes are sampled with replacement to maintain unbiased GCN embeddings. However, in practical implementations, sampling is often performed without replacement, introducing bias because the estimator is designed for sampling with replacement. A debiasing approach was introduced to improve accuracy and mitigate this bias. With \(s\) sampled indices, FastGCN/LADIES utilize the following importance sampling estimator to approximate the target sum of matrices:

\begin{align}
\frac{1}{\left | s_{node} \right |} \sum_{i \in [n]}  \frac{X_{s_{node_i}}}{ p_{s_{node_i}}}, \quad s_{node} = \left \{ 1, 2, \dots, s \right \}, \\ X = RP(HW).
\end{align}

This estimator represents a weighted average of \( RP(HW) \), and biases are introduced when the sampling is performed without replacement. To address this, the authors aim to preserve the linear form \( Y_s = \sum_{i=1}^{s} \beta_i X_{s_{\text{node}}} \) during the debiasing process and develop new coefficients \( \beta_i \) for each \( X_{s_{\text{node}}} \) to ensure that \( Y_s \) remains unbiased. Specifically, the debiasing is achieved through recursive weighted averaging as follows:

\begin{equation}
\begin{aligned}
\label{eq:calibProDeb}
Y_0 := 0, \quad Y_{i+1} = (1-\alpha_{i+1}) Y_i + \alpha_{i+1} \Pi_{i+1},\\
\text{where } \Pi_{i+1} = \sum_{j \in S_i} X_i + \frac{X_{s_{node_{i+1}}}}{p_{s_{node_{i+1}}}^i}, \quad \forall i = 0, 1, \dots, \left | s_{node} \right | -1.
\end{aligned}
\end{equation}

Here, \( S_i \) represents the set of indices sampled using weighted random sampling \cite{efraimidis2006weighted}, where \( 0 \leq i \leq |s_{node}| - 1 \) and \( S_0 = \emptyset \). $\alpha_1 = 1$, and $\alpha_{k+1}$ is a constant dependent on $k$. Specifically, $\alpha_{k+1} = \frac{n}{(n-k)(k+1)}$ is chosen to ensure that when all $p_i = 1/n$, the output coefficients align with those of a simple random sampling setting. This approach is referred to as 'debias' in the original paper. The LADIES and FastGCN variations incorporating this method are LADIES+debias (LADIES+d) and FastGCN+debias (FastGCN+d), respectively. Furthermore, sampling methods that incorporate both flat sampling \eqref{eq:calibProFlat} and debiasing \eqref{eq:calibProDeb} are referred to as LADIES+flat+debias (LADIES+f+d) and FastGCN+flat+debias (FastGCN+f+d), respectively.

\subsection{Feature Cache}\label{sec:cache} To accelerate training, Global Neighbor Sampling (GNS) \cite{dong2021global} facilitates efficient neighbor sampling within a mini-batch by periodically constructing a node cache, denoted as \( \mathcal{C} \), where \( \mathcal{C} = \{c_i \mid i \in \mathcal{V}_c \subset V\} \). The nodes in \( \mathcal{C} \) are selected using a biased sampling strategy to ensure they can be reached from the training set nodes with high probability. The features of nodes in \( \mathcal{C} \) are preloaded onto GPUs during training. To fit within GPU memory constraints, only 1\% of the nodes that are highly likely to be reachable from the training set are cached. This strategy effectively reduces data movement overhead and speeds up training by caching a small subset of high-degree nodes. GNS defines two approaches for determining the sampling probability of the cache. If the majority of the nodes in a graph belong to the training set, the sampling probability is determined based on the node degree. For a node $i$, the probability of being sampled in the cache is given by:

\begin{equation}
\label{eq:probLargeTrainNode}
p_{v_i} = \frac{\deg^{-}(v_i)}{\sum_{v_j \in V} \deg^{-}(v_j)},
\end{equation}

where \( \deg^{-}(v_i) \) denotes the in-degree of node $v_i$. In power-law graphs, caching a small subset of nodes is sufficient to cover most nodes due to the highly skewed degree distribution. When the training set includes only a small subset of the graph's nodes, GNS performs short random walks to compute the sampling probability. Considering sampled neighbors $\mathcal{N}_v$ of  $v \in V$, the sampling ratio \(\mathbf{d}\) is defined as:

\begin{equation}
\begin{aligned}
\label{eq:sampleProb}
\mathbf{d} = \left\{ \frac{\mathcal{N}(v_1)}{\deg^{-}(v_1)}, \frac{\mathcal{N}(v_2)}{\deg^{-}(v_2)}, \dots,  \frac{\mathcal{N}(v_{\left | V \right |})}{\deg^{-}(v_{\left | V \right |})} \right\}.
\end{aligned}
\end{equation}

The node sampling probability $P^{(l)} \in \mathbb{R}^{\left | V \right |}$ for the $l-th$ layer is computed as:

\begin{equation}
\begin{aligned}
\label{eq:probSmallTrain}
 P^{(l)} = \left( \mathbf{D} A + I \right) P^{(l-1)}, \quad \mathbf{D} = \operatorname{diag} \left( \mathbf{d} \right).\\
 p_{v_i}^{(0)} = \begin{cases}
  \frac{1}{\left | V_t \right |}, & \text{if } v_i \in V_t, \\ 
  0, & \text{otherwise}.
  \end{cases}
\end{aligned}
\end{equation}

where \(V_t\) denotes the training set. The final sampling probability for the cache is given by \( P^{(L)} \).

In summary, sampling techniques such as node-wise and layer-wise and their variants, such as flat and debiased LS, help mitigate computational and memory constraints in large-scale GNN training. These methods trade off scalability and accuracy while introducing challenges such as sampling bias and increased computational overhead. The node cache strategy further enhances efficiency by leveraging global neighbor sampling to preload frequently accessed nodes into GPU memory.

\Figure[t!](topskip=0pt, botskip=0pt, midskip=0pt)[width=1\linewidth]{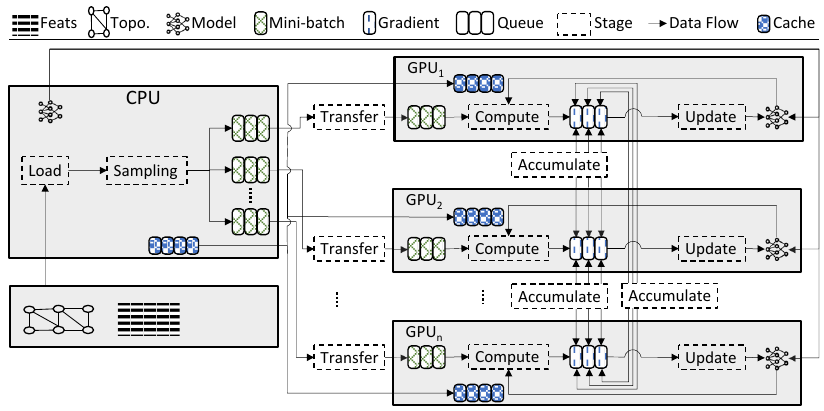}
{The MQ-GNN architecture integrates multi-queue pipelining to overlap mini-batch generation, data transfer, computation, asynchronous gradient sharing, and model updates across CPUs and GPUs. This design minimizes latency, maximizes GPU utilization, and accelerates large-scale GNN training.\label{fig:MQ-GNN Architecture}}

\section{Multi-Queue Pipelined GNN training Architecture}
This section presents the proposed MQ-GNN architecture for training GNN models. First, we provide an overview of the system design and a detailed analysis of each training stage.

\textbf{Overall Design} MQ-GNN introduces a pipelined architecture to optimize CPU, memory, and GPU utilization during GNN training. It employs mini-batch queues, gradient queues, and periodic synchronization to address scalability and efficiency challenges in multi-GPU environments. The pipeline architecture of MQ-GNN facilitates a smooth data flow between mini-batch generation, data transfer, and computation, as shown in Fig.~\ref{fig:MQ-GNN Architecture}. The architecture consists of seven stages: three dedicated to computation, three to data transfer, and one integrated stage combining computation and data transfer. The architecture uses multiple processes and threads across stages to maximize computational overlap and enable asynchronous processing. To efficiently manage data flow, we introduce three specialized queues: two for mini-batches and one for gradients. Each queue is populated using parallel processes and threads.

In the following, we provide the details of seven different stages of the MQ-GNN pipeline:

\textbf{Data Loading} In the initial stage, the system loads the graph's topology and features into the main memory, preparing the data for subsequent processing for GNN training.

\textbf{Model Transfer} Initially, the model is initialized and transferred to the GPU(s) for training. This ensures all GPUs begin with identical model parameters, enabling asynchronous processing across devices.

\textbf{Mini-Batch Generation and Enqueuing\label{sec:mini-bat_gen_enq}} In this stage, training batches are generated through sampling and enqueued into dedicated mini-batch queues on the CPU, with each GPU assigned its queue. This parallelized design reduces bottlenecks and enables efficient data transfer to GPUs.

Efficient mini-batch generation and management are crucial for improving training throughput, as Section~\ref{sec:prelimAndNotat} explains. To achieve this, MQ-GNN incorporates NS approaches, such as GCN \cite{kipf2016gcn} and GraphSAGE \cite{hamilton2017inductive}, as well as LS approaches, including FastGCN \cite{chen2018fastgcn}, LADIES \cite{zou2019layer}, and their advanced variants \cite{chen2023calibrate}, each discussed in Section~\ref{sec:prelimAndNotat}.

\textbf{Mini-Batch Transfer To GPU\label{sec:mini-batch_trans}} The generated mini-batches are transferred to the mini-batch queue on the GPU(s) during the \textit{Transfer} stage. Fig.~\ref{fig:MQ-GNN Architecture} illustrates the pipeline for mini-batch generation and transfer, highlighting how computation and data movement are overlapped. 

The CPU extracts features for nodes not present in the GPU cache. Missing features are gathered from the CPU's main memory and transferred to the GPU via PCIe, which can be resource-intensive. The overhead of this stage depends on the GPU cache size, cache policies, and graph properties like degree distribution. Efficient or complete GPU caching minimizes or eliminates this cost, optimizing data transfer and memory usage. Additionally, it eliminates the need for synchronization among GPUs when gathering neighboring features of target nodes on a GPU because each mini-batch on the GPU contains both features and topology components that are transferred from the CPU.

\textbf{Computation\label{sec:GPU_comp}} The \textit{compute} stage runs on the GPU(s), processing the enqueued mini-batches from the \textit{Transfer} stage. This stage focuses on processing mini-batches and performing forward (Fwd) and backward (Bwd) propagation to train the GNN concurrently with the previous two stages. The \textit{compute} stage focuses on maximizing GPU(s) utilization and optimizing compute operations by performing Fwd and Bwd operations without waiting for model updates from other GPUs. The worker thread consumes the mini-batches by performing Fwd propagation to train the GNN as follows:

\begin{equation} \label{eq:GNNFwd}
\begin{aligned}
Z^{l+1} \left (v \right )= GNN^l \left (\left [Z_{agg}^l \left(v \right),Z^l (v)  \right ], W^l \right), \\
Z_{agg}^l \left (v \right ) = Mean\left ( H^l \left ( v \right )\right ),
\end{aligned}
\end{equation}
where $H^l$ represents the intermediate embedding at layer $l$, $Z^l$ is the embedding at layer $l$, $W^l$ denotes the weight matrix, and $\left[ Z_{agg}^l (v), Z^l(v) \right]$ represents the concatenation of the aggregated neighborhood representation for GraphSAGE, and the previous embedding of node $v$ at layer $l$.

For GCN, $Z^{l+1} (v)$ is modified to replace the concatenation with direct neighborhood aggregation as follows:

\begin{equation}
Z^{l+1} (v) = GNN^l \left( Z_{agg}^l (v), W^l \right)
\end{equation}

As Fwd involves two main steps: gathering and aggregating neighboring features (Gather, GA) and applying the neural network (Apply NN, AN) \cite{ma2019neugraph, shao2024distributed}. From a matrix perspective, GA corresponds to  \( AH \) matrix multiplication, while AN corresponds to \( (AH)W \).  

The loss for a batch \( s \), \(\mathcal{L}^s\), is calculated as:  
\begin{eqnarray}
\mathcal{L}^s = - \sum_{v \in V^s} \left[ y_v \log(pr_v) + (1 - y_v) \log(1 - pr_v) \right],
\end{eqnarray}  
where \(V^s\) denotes the set of nodes in $s$, \( y_v \) denotes the ground truth label, and \( pr_v \) represents the predicted probability for node \( v \).

During backward propagation (Bwd), the gradient computation for the GA step, denoted as \(\triangledown\text{GA}\), involves \( AG \) operation, where \( G \) represents the gradient matrix. Neither GA nor \(\triangledown\text{GA}\) requires synchronization between GPUs.

After gradient computation, the gradients are enqueued into corresponding gradient queues rather than being synchronized immediately across GPUs. As illustrated in Fig.~\ref{fig:RACoM_fig}, GPU \( \mathcal{G}_0 \) enqueues the gradients during time \(\tau_1\) and proceeds to process another mini-batch, while the remaining GPUs (\( \mathcal{G}_i \), \( i \neq 0 \)) are still completing the first iteration. This pipelined execution facilitates optimal resource usage and overlapping operations.
In MQ-GNN, gradient queues store gradients awaiting synchronization, while mini-batch queues hold training data for GPU processing. This distinction ensures a clear demarcation between the pipeline's stages.

\textbf{Gradient Sharing and Accumulation Across GPUs\label{subsec:grad_share_acc}} 
 MQ-GNN employs asynchronous gradient-sharing mechanics, accumulating gradients across GPUs to balance updates. This process reduces the impact of communication delays by ensuring that local computations proceed without interruption. 

Once gradients are enqueued, they are asynchronously shared across all GPUs, ensuring prompt distribution without interrupting ongoing computations. Upon receiving delayed gradients, GPUs update their queues using the formula:

\begin{equation} \label{eq:grad_acc}
\Delta^{\text{New}} = \Delta^{\text{Current}} + \frac{\left(\Delta^{\mathcal{G}_i} - \Delta^{\text{Current}}\right)}{\text{Average Count}},
\end{equation}

where \( \Delta^{\mathcal{G}_i} \) is the incoming gradient from GPU $\mathcal{G}_i$, \( \Delta^{\text{Current}} \) is the existing gradient, and 'Average Count' tracks the number of gradients received (i.e., 'Average Count' \(\leq \left| \mathcal{G} \right|\)). This ensures balanced updates despite communication delays.

Fig.~\ref{fig:RACoM_fig} illustrates gradient sharing and accumulation, GPU \( \mathcal{G}_1\) enqueues and shares gradients after GPU \( \mathcal{G}_0\) during \(\tau_2\), while simultaneously processing data for the next iteration. This design ensures efficient execution, preventing delays caused by synchronization or resource under-utilization.

\Figure[t!](topskip=0pt, botskip=0pt, midskip=0pt)[width=1\linewidth]{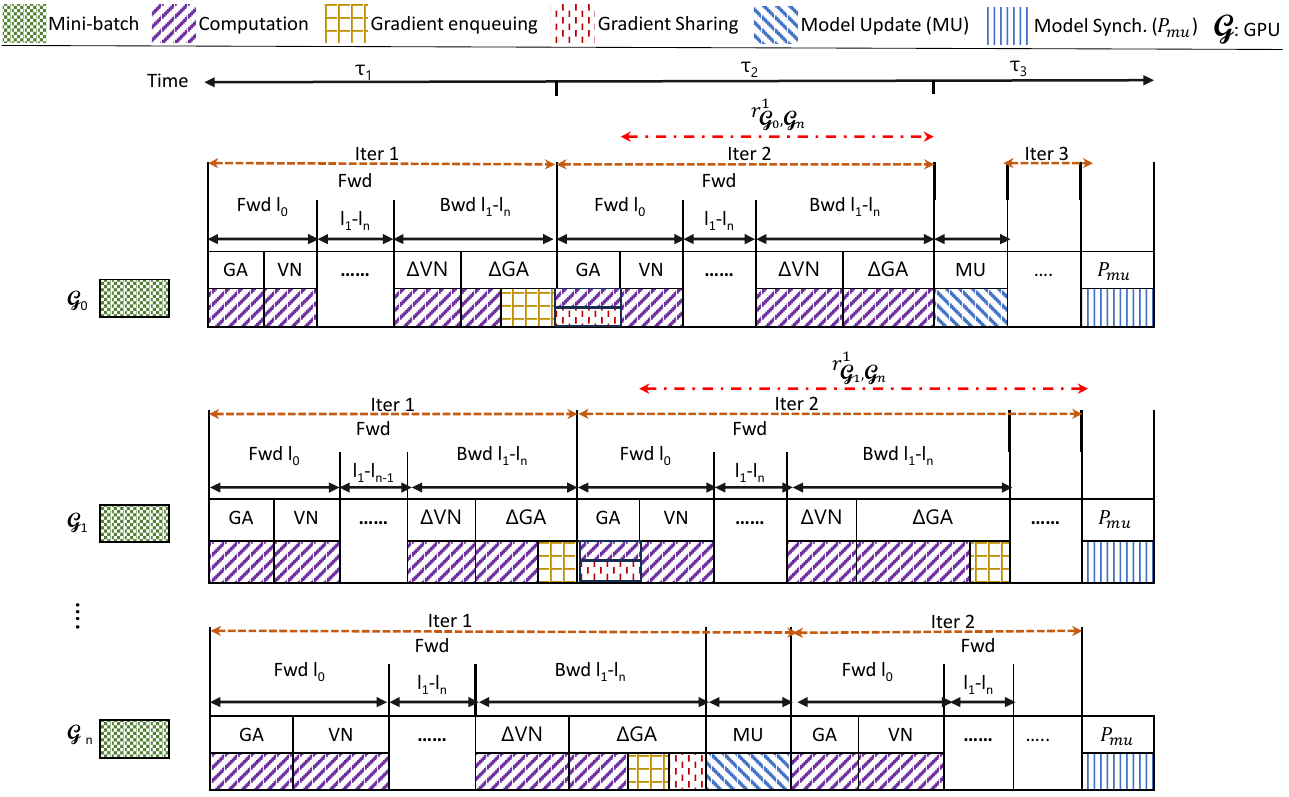}
{The RaCoM pipeline in MQ-GNN enables asynchronous gradient sharing, model updates, and periodic synchronization across GPUs. It efficiently overlaps mini-batch generation, data transfer, forward and backward propagation, and model synchronization \( P_{ms} \) to maximize resource utilization and minimize training latency.\label{fig:RACoM_fig}}

\textbf{Model Update on Each GPU\label{sec:model_update_gpus}} Once gradients from all GPUs are received (i.e., 'Average Count' \( = \left| \mathcal{G} \right|\)), the accumulated gradient is computed as follows:
\begin{equation} \label{eq:grad_acc_final}
 \Delta_l^{acc} = \frac{\sum_{i=1}^{\left| \mathcal{G} \right|} \Delta^{\mathcal{G}i}}{\left| \mathcal{G} \right|},
\end{equation}

where \( \Delta_l^{acc}\) is the accumulated gradient at layer \(l\), and \( \Delta_l^{\mathcal{G}i}\) is the batch-specific gradient computed at layer \(l\) on GPU \(\mathcal{G}_i\). This aggregation consolidates contributions from all GPUs processing different mini-batches. The \textit{Update} stage applies the accumulated gradients to update model parameters on GPUs.This stage gaurantees that model parameters are updated across all GPUs in a multi-GPU environment, integrating the collective contributions of the accumulated gradients. 

After completing the current iteration, each GPU incorporates the accumulated gradient into its model update before proceeding to the next iteration. The model update rule for layer $l$ is defined as:

\begin{equation}\label{eq:model_update}
W^l = W^l - \eta \Delta_l^{acc},
\end{equation}

 where $W^l$ denotes the weight matrix, and $\eta$ is the learning rate. Although gradient sharing facilitates efficient communication across GPUs, asynchronous updates can lead to model staleness. To mitigate this, we introduce the Ready-to-Update Asynchronous Consistent Model (RaCoM), a dynamic synchronization mechanism that ensures model consistency across GPUs.
 
\textbf{RaCoM\label{sec:racom}} RaCoM addresses the challenge of staleness in asynchronous GNN training by combining asynchronous local updates with periodic synchronization, ensuring global consistency and computational efficiency. The periodic synchronization mechanism adapts to the sparsity or density of the dataset, aligning updates to balance staleness and communication costs effectively. This design ensures that the GNN model achieves scalability while preserving accuracy, even under varying workload intensities. The synchronization period $P_{ms}$ is given by:

\begin{equation} \label{eq:peri_synch}
P_{ms} = \left\lceil \frac{\sqrt{|V|}}{\sqrt{\left| \mathcal{G} \right||E|}} \right\rceil,
\end{equation}

where \( |V| \) and \( |E| \) denote the number of nodes and edges in the graph, and \( \left| \mathcal{G} \right| \) is the number of GPUs. The synchronization period \( P_{\text{ms}} \) minimizes communication overhead while ensuring timely updates. This balance ensures that RaCoM can efficiently scale to handle large graphs and high GPU counts with a slight drop in model accuracy. In MQ-GNN, periodic synchronization, managed by \eqref{eq:peri_synch}, dynamically balances synchronization costs and staleness. MQ-GNN reduces the effect of asynchronous updates on training efficiency by adapting the synchronization period \( P_{\text{ms}} \) according to the structure of the graph. This ensures optimal utilization of computational resources across GPUs while maintaining model consistency. A detailed proof of the periodic synchronization mechanism and its derivation is provided in Appendix~\ref{app:periodic_sych}. These derivations highlight how \( P_{\text{ms}} \) dynamically adapts to graph properties to minimize synchronization costs.

Fig.~\ref{fig:RACoM_fig} shows the mechanism of the gradient-sharing mechanism and periodic synchronization proposed in MQ-GNN. It illustrates the gradient sharing process, highlighting how worker threads overlap gradient enqueuing and computation tasks. Gradient queues are used to streamline the communication of gradients between GPUs, enabling asynchronous updates and overlapping computation. It also depicts how periodic synchronization aligns gradients across GPUs to maintain consistency. Each gradient \( \Delta_t^{\mathcal{G}_i} \) generated by GPU \( \mathcal{G}_i \) at iteration \( t \) is first enqueued in its local gradient queue before being propagated to all other GPUs (\( \mathcal{G}_j \), \( j \neq i \)). The gradient \( \Delta_t^{\mathcal{G}_i} \) incurs a delay \( r_{\mathcal{G}_i, \mathcal{G}_j}^t \) before arriving at \( \mathcal{G}_j \). For instance, a straggler GPU (\( \mathcal{G}_n \)) shares its gradient \( \Delta_t^{\mathcal{G}_n} \) immediately after enqueuing, but the gradient reaches the other GPUs (\( \mathcal{G}_j, j \neq n \)) with a delay \( r_{\mathcal{G}_j, \mathcal{G}_n}^t \). When the gradients from the straggler GPU arrive at a receiving GPU, they are accumulated using \eqref{eq:grad_acc}. Once the gradients (i.e., \eqref{eq:grad_acc_final}) are ready to update the model, the receiving GPU completes the current iteration, incorporates the updated gradient into its model update \eqref{eq:model_update}, and proceeds to the next iteration. Finally, after \( P_{ms} \) iterations, model synchronization ensures consistency across all GPUs for maintaining model accuracy.

\subsection{Determining Optimal Queue Size}

When training GNNs on large-scale datasets, it is critical to identify the optimal queue size for efficient GPU utilization. A practical approach balances computational efficiency with memory constraints to ensure high throughput while avoiding overflows. The queue size is calculated based on batch processing times and the memory available on the GPU. This hybrid approach adjusts the queue size dynamically to maintain GPU utilization while respecting memory limits.

The queue size is given by:

\begin{equation} \label{eq:cal_queue_size}
\mathcal{Q} = \min\left(C, \max\left(2, \left\lceil \frac{\max(T_{\text{sampling}} + T_{\text{transfer}})}{\operatorname{mean}(T_{\text{compute}})} \right\rceil \right)\right) 
\end{equation}

Here, \( \max(T_{\text{sampling}} + T_{\text{transfer}}) \) represents the maximum time required for sampling and data transfer. In contrast, \( \operatorname{mean}(T_{\text{compute}}) \) denotes the average computation time on the GPU. The cap size, \( C \), sets an upper bound on the queue size to prevent exceeding the GPU’s memory capacity. The peak memory usage accounts for the temporary memory overheads incurred during CUDA computations. The $C$ is defined as:
\begin{equation} \label{eq:cal_cap_size}
C = \left\lfloor \frac{\mathcal{M}_{\mathcal{Q}}}{\mathcal{M}_{\text{mini-batch}}} \right\rfloor 
\end{equation}

where \( \mathcal{M}_{\mathcal{Q}} \) is the available memory for queues, and \( \mathcal{M}_{\text{mini-batch}} \) is the memory per mini-batch. The $\mathcal{M}_{\mathcal{Q}}$ is calculated by subtracting the measured peak memory usage (including a safety margin of 5–10\%) from the total GPU memory. The memory per batch depends on the batch size, feature dimension, and the data type used (e.g., 4 bytes for float32). This ensures that the queue size does not exceed the GPU’s memory capacity while maintaining efficiency.

Peak memory usage can be estimated by running a single training epoch using profiling tools such as PyTorch Profiler or NVIDIA Nsight Systems. These tools provide helpful information regarding memory allocation and the temporary overheads of CUDA computations. Based on this peak memory value, the peak capacity can be calculated and included in the queue size formula. 

\section{Evaluation}

This section comprehensively evaluates the MQ-GNN architecture by comparing it with baseline models.

\begin{table*}[ht] 
\caption{An overview of the datasets. One count is made for each undirected edge."Deg." stands for the graph's average degree. The ratio of training, validation, and test data is called the "split ratio."\label{tab:large-datasets}}
\begin{tabular}{llllllll}
\hline
Dataset       & Nodes     & Edges      & Avg. Degree & Feature dimension & Classes & Split Ratio & Metric   \\ \hline
ogbn-proteins & 132,534   & 39,561,252 & 597         & 8                 & 2       & 65/16/19         & ROC-AUC  \\
ogbn-arxiv    & 160,343   & 1,166,243  & 13.7        & 128               & 40      & 54/18/28         & Accuracy \\
Reddit        & 232965    & 11606919   & 50          & 602               & 41      & 66/10/24          & F1-score \\
ogbn-products & 2,449,029 & 61,859,140 & 50.5        & 100               & 47      & 8/2/90           & Accuracy \\ \hline
\end{tabular}
\end{table*}

\subsection{Experimental Setup}

This section describes the hardware setup used for the experiments.

\textbf{Hardware Setup} The experiments were conducted on Linux clusters managed by the SLURM workload manager. The hardware setup in this work was determined by the maximum resource allocation permitted by the SLURM cluster account. Each job was allocated up to 4 GPUs, 32 CPUs, and 126 GB of RAM, with a maximum runtime of 4 days. 

The experiments were conducted on a machine running Ubuntu 20.04.6 LTS (kernel 5.4.0-192-generic), equipped with four NVIDIA GeForce RTX 3090 GPUs (24 GB each) and 32 Intel Xeon E-2234 CPUs (3.60 GHz, four cores per socket, two threads per core). To maximize these resources, we optimized CPU, GPU, and memory use across single, two, three, and four GPU setups.

\begin{itemize}
    \item \textbf{Experiments with a Single GPU} Each experiment used 32 CPUs and 128 GB of memory.
    \item \textbf{Experiments with Two GPUs} Each experiment used 16 CPUs per GPU and 64 GB of memory per GPU.
    \item \textbf{Experiments with Three or Four GPUs} Each experiment used 8 CPUs per GPU and 32 GB of memory per GPU.
\end{itemize}

This scaling across GPU configurations enables the evaluation of MQ-GNN's performance and scalability under different parallelism and resource contention levels.

\textbf{Datasets and Evaluation Metrics} To ensure fair comparisons and representative results, we conducted experiments on four widely used benchmark datasets: Reddit \cite{hamilton2017inductive} and three Open Graph Benchmark (OGB) datasets—ogbn-arxiv, ogbn-proteins, and ogbn-products \cite{hu2020Open}. Table~\ref{tab:large-datasets} provides a summary of these datasets. These datasets exhibit diverse properties. The Reddit dataset, widely used in prior works \cite{chen2018fastgcn, chen2018stochastic}, contains high-degree nodes and complex community structures, making it a standard benchmark for evaluating scalability. This dataset presents challenges due to its high average node degree, large feature dimensions, and dense subgraphs, necessitating efficient sampling strategies. The OGB datasets present additional challenges, including a class imbalance in ogbn-proteins and many nodes in ogbn-products, which test both scalability and accuracy \cite{hu2020Open}. Together, these datasets provide a comprehensive testbed for evaluating the scalability, efficiency, and generalization capabilities of MQ-GNN across diverse graph structures.

We used different evaluation metrics for each dataset, as shown in Table~\ref{tab:large-datasets}, maintaining consistency with recent literature and following the recommendations of \cite{hu2020Open} to ensure compatibility with baselines.

\textbf{Models and Hyperparameters} To ensure a fair comparison, all baseline models were implemented using PyTorch and DGL, maintaining consistent training parameters across all methods. For node-wise sampling, five neighbors were chosen per node, while for layer-wise sampling, 512 nodes were selected per layer. All models were trained using a two-layer GNN with the ADAM optimizer, set to a learning rate of \( \eta = 0.001 \). A caching strategy was employed, periodically storing \(1\%\) of the nodes per epoch to enhance efficiency. Early stopping was applied if the validation metric did not improve by at least 0.01 for 200 consecutive batches. The model with the highest validation metric was selected as the final model. Each model was trained ten times per method and dataset to ensure statistical robustness, with the mean and variance reported for evaluation. To mitigate GPU timing variability caused by kernel initialization and synchronization overheads, the first and last 20 mini-batch timings were excluded. This ensures the reported timings reflect steady-state performance, reducing noise from warm-up and cool-down phases.

\subsection{Comparison with Existing Systems}
To demonstrate that MQ-GNN optimizes resource utilization more effectively than existing systems, leading to faster training, we compare it against a range of baseline models. These include NS-based methods such as mini-batch GCN and GraphSAGE \cite{wang2019dgl}, as well as LS-based methods, including FastGCN \cite{chen2018fastgcn}, LADIES, and their advanced variants \cite{chen2023calibrate}. These baselines were selected based on their prominence in prior research and effectiveness in representing state-of-the-art NS-based and LS-based GNN training approaches.

The evaluation results, summarized in Tables~\ref{tab:g1-ns-results} to~\ref{tab:g4-ladies-large-results}, report the mean ($\pm$ standard deviation) of batch and training times (in milliseconds, ms), along with evaluation metric percentages for each dataset. Table \ref{tab:large-datasets} summarizes dataset characteristics. 

We first highlight key insights from the results before presenting a detailed discussion.

\textbf{Major Highlights} The MQ-GNN architecture significantly improves performance by optimizing resource utilization and reducing overhead in GNN training. Across multiple datasets, MQ-GNN achieves up to $4.6\,\times$ faster training speeds than baseline models while maintaining comparable evaluation metrics. This acceleration is primarily due to the MQ-GNN's efficient multi-queue design, which enables concurrent computation, data transfer, gradient sharing, and periodic model synchronization across GPUs. Moreover, MQ-GNN adapts effectively to various datasets and configurations, consistently achieving performance gains across various GPU setups, sampling strategies, and node cache sizes. These results underscore MQ-GNN's scalability, efficiency, and effectiveness in large-scale GNN training.

\begin{table*}[!ht] 
\caption{Results for batch time (ms), training time (ms), and evaluation metrics (\%) for GCN, GraphSAGE, MQ-GCN, and MQ-GraphSAGE on a single-GPU configuration.\label{tab:g1-ns-results}}
\begin{tabular}{llllll}
\hline
                               & Benchmark          & ogbn-proteins           & ogbn-arxiv                & Reddit                    & ogbn-products            \\ \hline
\multirow{3}{*}{GCN}       & Batch              & 10.29 $\pm$ 0.20        & 11.53 $\pm$ 0.15          & 24.70 $\pm$ 0.18          & 14.11 $\pm$ 0.24          \\
                                & Training           & 351669.12 $\pm$ 3616.96 & 505270.1 $\pm$ 4952.69    & \textbf{1471507.1 $\pm$ 15604.08}  & 1081781.22 $\pm$ 21635.62 \\
                                & Metrics            & 66.31 $\pm$ 0.50        & 67.69 $\pm$ 0.55          & 88.81 $\pm$ 0.80          & 76.10 $\pm$ 0.48          \\ \hline
\multirow{3}{*}{MQ-GCN}         & Batch              & \textbf{6.43 $\pm$ 0.15}         & \textbf{6.50 $\pm$ 0.20}           & 11.89 $\pm$ 0.18          & 6.83 $\pm$ 0.22           \\
                                & Training           & \textbf{197371.73 $\pm$ 1953.72} & \textbf{280705.61 $\pm$ 2807.06}   & \textbf{731104.99 $\pm$ 8175.05}   & 559839.94 $\pm$ 6011.34   \\
                                & Metrics            & 66.87 $\pm$ 0.35        & 67.25 $\pm$ 0.40          & 89.71 $\pm$ 0.55          & 76.95 $\pm$ 0.40          \\ \hline
\multirow{3}{*}{GraphSAGE} & Batch              & 23.03 $\pm$ 0.25        & 28.11 $\pm$ 0.30          & 38.01 $\pm$ 0.35          & 21.23 $\pm$ 0.40          \\
                                & Training           & 786946.56 $\pm$ 6993.38 & 1050783.26 $\pm$ 12418.31 & 2326703.17 $\pm$ 22372.31 & \textbf{1725299.42 $\pm$ 25712.47} \\
                                & Metrics            & 66.01 $\pm$ 0.45        & 68.51 $\pm$ 0.55          & 94.36 $\pm$ 0.25          & 76.87 $\pm$ 0.65          \\ \hline
\multirow{3}{*}{MQ-GraphSAGE}   & Batch              & 15.99 $\pm$ 0.20        & 17.99 $\pm$ 0.25          & 23.36 $\pm$ 0.30          & 12.27 $\pm$ 0.25          \\
                                & Training           & 536981.42 $\pm$ 4360.81 & 639797.48 $\pm$ 5837.68   & 1351612.87 $\pm$ 12926.13 & \textbf{958494.12 $\pm$ 9584.94}   \\
                                & Metrics            & 66.11 $\pm$ 0.35        & 68.41 $\pm$ 0.40          & 94.41 $\pm$ 0.30          & 77.01 $\pm$ 0.40          \\ \hline
                                
\end{tabular}
\end{table*}

\subsection{Results and Discussion}
This section evaluates MQ-GNN's performance across various GNN models and standard datasets. Specifically, we analyze its performance on both single-GPU and multi-GPU configurations, leveraging NS and LS sampling approaches.

\subsubsection{Single GPU Performance}

MQ-GNN demonstrates substantial performance gains in single-GPU configurations, achieving significantly faster training and batch processing times than baselines. MQ-GNN reduces training time by up to \(2.06\,\times\) across four datasets while maintaining comparable evaluation metrics.

\textbf{NS Performance} NS methods such as MQ-GraphSAGE demonstrate significant performance improvements. For Reddit, with its dense graph structure and high-degree nodes, MQ-GCN achieved a \(2.01\,\times\) speedup, as shown in Table \ref{tab:g1-ns-results}, demonstrating the efficiency of MQ-GNN’s queuing mechanism in managing data transfer and computation for large feature sizes and dense neighborhoods. Similarly, on the ogbn-products dataset, MQ-GraphSAGE reduced training time by \(1.81\,\times\) and improved batch time by \(1.98\,\times\). MQ-GNN achieves significant performance gains on the Reddit dataset due to its dense structure and large feature dimensions. The significant feature dimensions and dense connectivity enable the queuing mechanism to interleave data transfer and computation, improving GPU utilization efficiently. In contrast, the dense structure of ogbn-products facilitates the queuing mechanism, but its large graph size increases data transfer demands. Although the smaller feature size allows for some caching benefits, it is less impactful than Reddit, which has larger feature dimensions.

MQ-GCN maintains efficiency even on smaller graphs such as ogbn-proteins and sparse datasets such as ogbn-arxiv. Table \ref{tab:g1-ns-results} shows that MQ-GCN achieved a \(1.5\,\times\) speedup on ogbn-arxiv by leveraging its moderate graph density and larger feature dimension (128). Due to the more prominent feature dimension, MQ-GNN’s caching approach improves GPU efficiency by reducing redundant data transfers and effectively interleaving data transfer with GNN computation. In contrast, MQ-GCN reduced training time by \(1.4\,\times\) on the ogbn-proteins dataset. The smaller feature dimension (8) limits the effectiveness of caching and reduces the efficiency of interleaving data transfer with GNN computation, resulting in comparatively more minor gains than on ogbn-arxiv.

\begin{table*}[t] 
\caption{Results for batch time (ms), training time (ms), and evaluation metrics (\%) for  FastGCN, MQ-FastGCN, and their enhanced variants on a single-GPU configuration.\label{tab:g1-fastgcn-large-results}}
\begin{tabular}{llllll}
\hline
Model                             & Benchmark & ogbn-proteins           & ogbn-arxiv              & Reddit                  & ogbn-products            \\ \hline
\multirow{3}{*}{FastGCN}     & Batch     & 8.99 $\pm$ 0.20         & 8.84 $\pm$ 0.25         & 9.75 $\pm$ 0.30         & 10.09 $\pm$ 0.50         \\
                                  & Training  & 307300.5 $\pm$ 3073.01  & 316598.2 $\pm$ 26256.19 & 588347.7 $\pm$ 4975.32  & 400561.65 $\pm$ 7831.35  \\
                                  & Metrics   & 53.2 $\pm$ 0.50         & 26.11 $\pm$ 0.40        & 45.7 $\pm$ 0.55         & 27.3 $\pm$ 0.55          \\ \hline
\multirow{3}{*}{MQ-FastGCN}       & Batch     & 5.42 $\pm$ 0.18         & 5.11 $\pm$ 0.20         & 5.39 $\pm$ 0.22         & 5.01 $\pm$ 0.25          \\
                                  & Training  & 180722.50 $\pm$ 1707.23 & 173888.99 $\pm$ 1758.89 & 298911.13 $\pm$ 3268.60 & 200101.12 $\pm$ 2225.34  \\
                                  & Metrics   & 53.1 $\pm$ 0.45         & 26.21 $\pm$ 0.40        & 45.9 $\pm$ 0.50         & 27.1 $\pm$ 0.50          \\ \hline
\multirow{3}{*}{FastGCN+f}   & Batch     & 8.57 $\pm$ 0.15         & 8.85 $\pm$ 0.18         & 9.18 $\pm$ 0.20         & 10.07 $\pm$ 0.45         \\
                                  & Training  & 292837.5 $\pm$ 2839.29  & 316811.5 $\pm$ 3057.12  & 553561.12 $\pm$ 5535.61 & 390916.17 $\pm$ 6929.41  \\
                                  & Metrics   & 61.1 $\pm$ 0.45         & 23.61 $\pm$ 0.35        & 53.61 $\pm$ 0.49        & 33.22 $\pm$ 0.51         \\ \hline
\multirow{3}{*}{MQ-FastGCN+f}     & Batch     & 5.36 $\pm$ 0.15         & 5.23 $\pm$ 0.18         & 5.74 $\pm$ 0.20         & 6.29 $\pm$ 0.22          \\
                                  & Training  & 162687.50 $\pm$ 1626.88 & 176006.39 $\pm$ 1760.06 & 307534.00 $\pm$ 3075.34 & 217175.65 $\pm$ 2171.76  \\
                                  & Metrics   & 61.12 $\pm$ 0.30        & 23.13 $\pm$ 0.25        & 54.11 $\pm$ 0.40        & 33.21 $\pm$ 0.35         \\ \hline
\multirow{3}{*}{FastGCN+d}   & Batch     & 8.52 $\pm$ 0.15         & 8.65 $\pm$ 0.18         & 9.31 $\pm$ 0.25         & 9.97 $\pm$ 0.50          \\
                                  & Training  & 291145.5 $\pm$ 2844.56  & 309764.5 $\pm$ 3284.55  & 561751.15 $\pm$ 4727.52 & 386924.19 $\pm$ 6736.37  \\
                                  & Metrics   & 53.69 $\pm$ 0.51        & 27.32 $\pm$ 0.65        & 44.25 $\pm$ 0.70        & 27.97 $\pm$ 0.65         \\ \hline
\multirow{3}{*}{MQ-FastGCN+d}     & Batch     & 5.23 $\pm$ 0.18         & 5.01 $\pm$ 0.20         & 5.12 $\pm$ 0.22         & 5.23 $\pm$ 0.25          \\
                                  & Training  & 181747.50 $\pm$ 1617.48 & 171091.39 $\pm$ 1720.91 & 310028.42 $\pm$ 3110.28 & 202957.88 $\pm$ 2149.58  \\
                                  & Metrics   & 53.75 $\pm$ 0.51        & 27.25 $\pm$ 0.65        & 44.66 $\pm$ 0.70        & 27.84 $\pm$ 0.65         \\ \hline
\multirow{3}{*}{FastGCN+f+d} & Batch     & 8.65 $\pm$ 0.20         & 8.42 $\pm$ 0.18         & 9.47 $\pm$ 0.22         & 20.22 $\pm$ 0.60         \\
                                  & Training  & \textbf{295582.1 $\pm$ 3023.79}  & 301607.1 $\pm$ 3161.71  & 571206.23 $\pm$ 4821.12 & \textbf{784496.26 $\pm$ 15689.93} \\
                                  & Metrics   & 61.1 $\pm$ 0.75         & 24.79 $\pm$ 0.70        & 53.97 $\pm$ 0.85        & 30.11 $\pm$ 1.50         \\ \hline
\multirow{3}{*}{MQ-FastGCN+f+d}   & Batch     & 5.39 $\pm$ 0.20         & 5.27 $\pm$ 0.22         & 5.94 $\pm$ 0.25         & 6.12 $\pm$ 0.30          \\
                                  & Training  & \textbf{163400.35 $\pm$ 1634.00} & 170154.30 $\pm$ 1701.54 & 297258.84 $\pm$ 2972.59 & \textbf{378999.76 $\pm$ 3790.00}  \\
                                  & Metrics   & 61.13 $\pm$ 0.85        & 24.85 $\pm$ 0.70        & 53.99 $\pm$ 0.85        & 29.97 $\pm$ 1.50         \\ \hline
\end{tabular}
\end{table*}

\begin{table*}[] 
\caption{Results for batch time (ms), training time (ms), and evaluation metrics (\%) for  LADIES, MQ-LADIES, and their enhanced variants on a single-GPU configuration.\label{tab:g1-ladies-large-results}}
\begin{tabular}{llllll}
\hline
Model                            & Benchmark & ogbn-proteins           & ogbn-arxiv              & Reddit                    & ogbn-products            \\ \hline
\multirow{3}{*}{LADIES}     & Batch     & 10.18 $\pm$ 0.20        & 11.39 $\pm$ 0.15        & 13.26 $\pm$ 0.18          & 9.11 $\pm$ 0.40          \\
                                 & Training  & 348144.5 $\pm$ 4021.12  & \textbf{604201.54 $\pm$ 5898.82} & 807788.34 $\pm$ 7965.98   & 476977.05 $\pm$ 9539.54  \\
                                 & Metrics   & 68.27 $\pm$ 0.12        & 61.0 $\pm$ 0.20         & 74.9 $\pm$ 0.18           & 52.84 $\pm$ 0.40         \\ \hline
\multirow{3}{*}{MQ-LADIES}       & Batch     & 6.51 $\pm$ 0.18         & 7.11 $\pm$ 0.20         & 7.64 $\pm$ 0.18           & 4.99 $\pm$ 0.25          \\
                                 & Training  & 211709.12 $\pm$ 2427.09 & \textbf{348106.70 $\pm$ 2851.07} & 442648.37 $\pm$ 5726.48   & 248904.21 $\pm$ 2359.04  \\
                                 & Metrics   & 68.37 $\pm$ 0.10        & 61.4 $\pm$ 0.15         & 75.0 $\pm$ 0.20           & 52.71 $\pm$ 0.25         \\ \hline
\multirow{3}{*}{LADIES+f}   & Batch     & 10.24 $\pm$ 0.22        & 11.27 $\pm$ 0.18        & 13.29 $\pm$ 0.20          & 9.13 $\pm$ 0.50          \\
                                 & Training  & 349983.5 $\pm$ 2799.98  & 493595.2 $\pm$ 5365.12  & 805818.11 $\pm$ 7933.98   & 528674.09 $\pm$ 10573.48 \\
                                 & Metrics   & 68.18 $\pm$ 0.10        & 62.8 $\pm$ 0.18         & 90.12 $\pm$ 0.20          & 62.14 $\pm$ 0.30         \\ \hline
\multirow{3}{*}{MQ-LADIES+f}     & Batch     & 6.35 $\pm$ 0.18         & 6.61 $\pm$ 0.22         & 7.36 $\pm$ 0.18           & 4.79 $\pm$ 0.25          \\
                                 & Training  & 201255.32 $\pm$ 2432.55 & 269853.61 $\pm$ 2858.54 & 416927.39 $\pm$ 5769.27   & 247821.93 $\pm$ 2378.22  \\
                                 & Metrics   & 68.38 $\pm$ 0.15        & 62.82 $\pm$ 0.20        & 90.21 $\pm$ 0.25          & 62.11 $\pm$ 0.40         \\ \hline
\multirow{3}{*}{LADIES+d}   & Batch     & 11.62 $\pm$ 0.25        & 13.0 $\pm$ 0.22         & 14.28 $\pm$ 0.25          & 9.18 $\pm$ 0.60          \\ 
                                 & Training  & 397178.5 $\pm$ 2989.76  & 571631.9 $\pm$ 4979.89  & 1063068.25 $\pm$ 11520.23 & 432755.37 $\pm$ 8655.11  \\
                                 & Metrics   & 68.76 $\pm$ 0.15        & 61.48 $\pm$ 0.22        & 86.81 $\pm$ 0.25          & 54.82 $\pm$ 0.50         \\ \hline
\multirow{3}{*}{MQ-LADIES+d}     & Batch     & 7.31 $\pm$ 0.25         & 7.98 $\pm$ 0.22         & 8.40 $\pm$ 0.25           & 5.09 $\pm$ 0.30          \\
                                 & Training  & 248188.67 $\pm$ 2521.89 & 334029.90 $\pm$ 2990.30 & 590965.54 $\pm$ 5929.66   & 239021.27 $\pm$ 2510.21  \\
                                 & Metrics   & 68.9 $\pm$ 0.20         & 61.45 $\pm$ 0.25        & 86.89 $\pm$ 0.30          & 54.98 $\pm$ 0.35         \\ \hline
\multirow{3}{*}{LADIES+f+d} & Batch     & 11.58 $\pm$ 0.22        & \textbf{14.96 $\pm$ 0.25}        & 15.35 $\pm$ 0.30          & 9.11 $\pm$ 0.70          \\
                                 & Training  & 395726.1 $\pm$ 2989.67  & 697795.4 $\pm$ 5969.87  & 1381972.57 $\pm$ 14728.65 & 404781.39 $\pm$ 15895.63 \\
                                 & Metrics   & 67.55 $\pm$ 0.20        & 61.90 $\pm$ 0.25        & 88.43 $\pm$ 0.30          & 62.20 $\pm$ 0.60         \\ \hline
\multirow{3}{*}{MQ-LADIES+f+d}   & Batch     & 7.09 $\pm$ 0.22         & 7.55 $\pm$ 0.25         & 8.45 $\pm$ 0.30           & 5.01 $\pm$ 0.30          \\
                                 & Training  & 231872.88 $\pm$ 2518.73 & 383751.93 $\pm$ 2997.52 & 694872.38 $\pm$ 5948.72   & 201580.35 $\pm$ 2515.80  \\
                                 & Metrics   & 67.49 $\pm$ 0.25        & 62.01 $\pm$ 0.30        & 88.29 $\pm$ 0.35          & 62.30 $\pm$ 0.40         \\ \hline
\end{tabular}
\end{table*}

\textbf{LS Performance} LS methods benefited greatly from MQ-GNN's pipelining mechanism. For ogbn-products, MQ-FastGCN+f+d achieved a \(2.06\,\times\) reduction in training time (Table~\ref{tab:g1-fastgcn-large-results}), demonstrating its ability to manage the dataset's large scale and high feature dimensionality using multi-queue and caching mechanisms. For smaller graphs like ogbn-proteins, MQ-FastGCN+f+d reduced training time by \(1.6\,\times\), while MQ-LADIES+f+d dropped it by \(1.7\,\times\) (Tables~\ref{tab:g1-fastgcn-large-results} and~\ref{tab:g1-ladies-large-results}), demonstrating MQ-GNN's versatility across varying dataset scales and graph structures. MQ-GNN's queuing method reduces GPU idle times and improves utilization by increasing the overlap between computation and data transfer. Mini-batch preparation is accelerated by the caching mechanism's ability to reduce redundant node access due to the dense graph connectivity.

\begin{figure}[t]
 \centering
 \includegraphics[width=.5\textwidth]{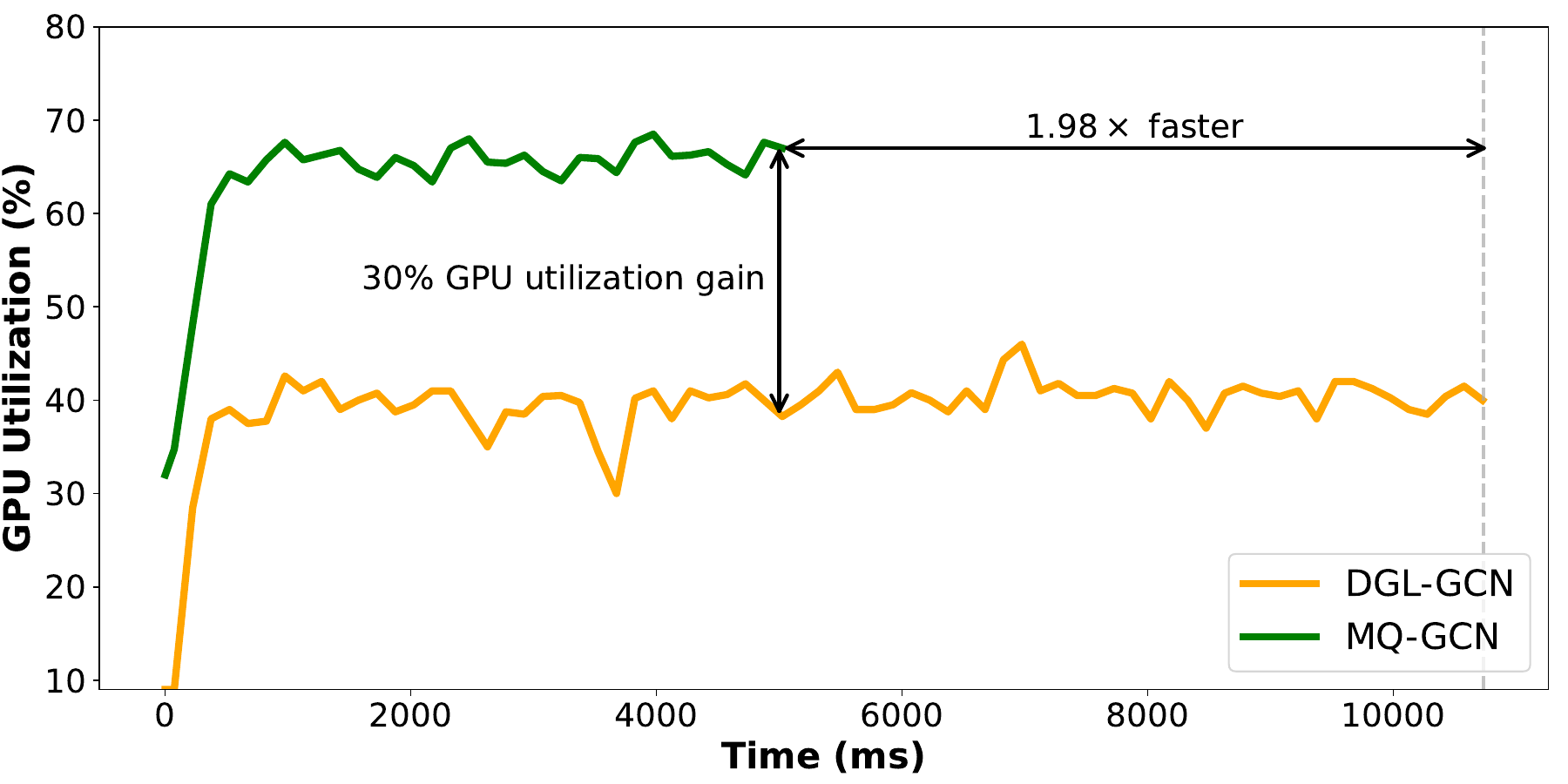}
 \caption{The GPU utilization of MQ-GCN and DGL-GCN during a single epoch of training on Reddit. Utilization is smoothed over 150 ms.\label{fig:gpu_util}}
\end{figure}

\textbf{Comparative Analysis} Across all datasets and LS and NS strategies, MQ-GNN consistently outperformed baselines, achieving faster training times and batch processing speeds without compromising metrics. Separating sampling from training eliminates bottlenecks observed in traditional systems like LADIES, where sampling distribution updates (Section~\ref{sec:ladies}) slow down mini-batch preparation. MQ-GNN's queuing mechanism overlaps data transfer, sampling, and computation phases to ensure continuous GPU utilization. Additionally, MQ-GNN effectively integrates processes with higher computational costs, such as debiasing \eqref{eq:calibProDeb} and flat sampling \eqref{eq:calibProFlat} in LADIES+f+d and FastGCN+f+d, into its pipeline to minimize its runtime impact.

We also compare GPU utilization during training on a single epoch of the Reddit dataset using MQ-GCN and GCN, as shown in Fig.~\ref{fig:gpu_util}, where MQ-GNN achieves a \(1.98\,\times\) speedup and a \(30\%\) increase in GPU utilization. MQ-GNN achieves significantly higher GPU utilization, maintaining an average utilization of approximately 64.2\% with a peak utilization of 73\%, compared to DGL, which stabilizes around 39.42\%. A key factor influencing GPU utilization is determining the optimal queue size and interleaving data transfer and GNN computation to reduce GPU starvation for data. 

In MQ-GNN, the queue size directly affects the overlap between batch preparation and GPU computation. Our analysis shows that datasets with higher sampling and data transfer times, such as Reddit, required larger queue sizes to utilize the GPU fully. Conversely, smaller datasets like ogbn-arxiv, with lower variability in batch preparation times, achieved better performance with smaller queue sizes. 

MQ-GNN dynamically adjusted queue sizes to maximize throughput while avoiding memory overflows by profiling workload characteristics and applying a formula  \eqref{eq:cal_queue_size} that accounts for maximum preparation time, average compute time and memory constraints. For example, in the Reddit dataset, where the peak memory usage was 730.95 MB and the maximum combined sampling and data transfer time was 51 ms, a queue size of 5 ensured steady GPU utilization without exceeding memory limits. Similarly, in the ogbn-products dataset, where peak memory usage was 240.62 MB and preparation times were lower, a smaller queue size of 3 was sufficient to maintain optimized GPU utilization. 

These results highlight the adaptability of MQ-GNN in addressing dataset-specific challenges and ensuring efficient memory usage and consistent GPU utilization. MQ-GNN demonstrated higher peak GPU utilization and more excellent stability than DGL, which exhibited frequent dips in utilization, likely caused by sequential data processing and transfer operations that introduced delays. This stability results from integrating queuing mechanisms and memory-based constraints that optimize data pipelines. 

In addition, MQ-GNN prevents GPU memory saturation for large datasets by limiting queue sizes according to available GPU memory and batch size requirements. MQ-GNN's profile-driven queue sizing achieves better utilization and balances resource constraints and computational efficacy. This method involves profiling peak memory usage per dataset and predicting queue sizes under GPU memory limits, considering temporary data utilized while performing CUDA operations. This approach enhances GPU utilization. These results demonstrate the advantages of MQ-GNN's queuing mechanism and asynchronous data handling, which sustain GPU activity and enhance overall throughput.

MQ-GNN's lightweight design eliminates the gradient queue for single-GPU configurations, simplifies the training pipeline, and reduces training time. Its queuing and pipelining mechanisms enable these improvements by decoupling sampling and computation, increasing the overlap between data transfer and training. By avoiding sequential processing, where GPUs are often idle during data preparation, MQ-GNN eliminates these bottlenecks, resulting in consistent GPU utilization and improved throughput. These findings establish MQ-GNN as a robust, scalable, and efficient solution for GNN training, even in single-GPU setups.

\begin{table*}[] 
\caption{Results for batch time (ms), training time (ms), and evaluation metrics (\%) for GCN, GraphSAGE, MQ-GCN, and MQ-GraphSAGE on a two-GPU configuration.\label{tab:g2-ns-results}}
\begin{tabular}{llllll}
\hline
Model                           & Benchmark & ogbn-proteins            & ogbn-arxiv              & Reddit                    & ogbn-products           \\ \hline
\multirow{3}{*}{GCN}        & Batch     & 12.15 $\pm$ 0.58         & 13.84 $\pm$ 0.47        & \textbf{23.54 $\pm$ 0.42}          & 13.32 $\pm$ 0.49        \\
                                & Training  & 207644.09 $\pm$ 11074.32 & 298242.34 $\pm$ 3345.21 & \textbf{716904.12 $\pm$ 7458.76}   & 224901.25 $\pm$ 5289.03 \\
                                & Metrics   & 66.28 $\pm$ 0.49         & 67.71 $\pm$ 0.52        & \textbf{88.35 $\pm$ 0.79}          & 75.36 $\pm$ 0.51        \\ \hline
\multirow{3}{*}{MQ-GCN}         & Batch     & 4.81 $\pm$ 8.25          & 5.20 $\pm$ 9.12         & \textbf{7.47 $\pm$ 22.50}          & 4.61 $\pm$ 10.34        \\
                                & Training  & 83392.37 $\pm$ 0.49      & 114318.65 $\pm$ 0.52    & \textbf{230592.43 $\pm$ 0.79}      & 80000.44 $\pm$ 0.51     \\
                                & Metrics   & 66.08 $\pm$ 0.40         & 67.61 $\pm$ 0.45        & \textbf{88.3 $\pm$ 0.70}           & 75.26 $\pm$ 0.40        \\ \hline
\multirow{3}{*}{GraphSAGE} & Batch     & 23.71 $\pm$ 0.40         & 30.85 $\pm$ 0.61        & 36.13 $\pm$ 0.55          & 23.09 $\pm$ 0.37        \\
                                & Training  & \textbf{405154.41 $\pm$ 4358.29}  & 637116.36 $\pm$ 6728.41 & \textbf{1132782.14 $\pm$ 11834.12} & 504154.02 $\pm$ 5310.89 \\
                                & Metrics   & \textbf{66.03 $\pm$ 0.42}         & \textbf{68.58 $\pm$ 0.57}        & 94.89 $\pm$ 0.23          & \textbf{78.39 $\pm$ 0.62}        \\ \hline
\multirow{3}{*}{MQ-GraphSAGE}   & Batch     & 8.91 $\pm$ 12.51         & 11.41 $\pm$ 15.34       & 11.21 $\pm$ 27.56         & 7.62 $\pm$ 13.28        \\
                                & Training  & \textbf{154708.64 $\pm$ 0.50}     & 241777.02 $\pm$ 0.52    & \textbf{359521.25 $\pm$ 0.80}      & 171465.83 $\pm$ 0.62    \\
                                & Metrics   & \textbf{65.73 $\pm$ 0.35}         & \textbf{68.38 $\pm$ 0.40}        & 94.84 $\pm$ 0.18          & \textbf{78.19 $\pm$ 0.50}        \\ \hline
\end{tabular}
\end{table*}

\subsubsection{Multi-GPU Performance}
MQ-GNN provides significant performance gains in multi-GPU configurations compared to baselines (up to \(4.6\,\times\)), as evident from Tables~\ref{tab:g2-ns-results}-\ref{tab:g4-ladies-large-results}. By leveraging the RaCoM framework (Section~\ref{sec:racom}), gradient sharing with queues, and efficient queuing mechanisms, MQ-GNN demonstrates faster training and batch times across diverse datasets while maintaining comparable evaluation performance despite a drop of approximately $2.0\%$ in evaluation metrics. These results highlight the system's ability to reduce communication overhead, minimize GPU idle time, and handle staleness issues through periodic synchronization.

\textbf{NS Performance}
In a two-GPU configuration, MQ-GNN achieves substantial reductions in training and batch times while maintaining competitive metrics, as shown in Tables~\ref{tab:g2-ns-results}-\ref{tab:g2-ladies-large-results}. For example, MQ-GCN significantly improved over GCN on the Reddit dataset, with a training time reduction of \(3.10\,\times\). MQ-GraphSAGE also achieved a \(3.15\,\times\) speedup (Table~\ref{tab:g2-ns-results}). These results highlight MQ-GNN's efficient queuing mechanism, which reduces GPU idle time and enables smooth mini-batch transitions. Despite these improvements, there was a minor decline in metrics, with MQ-GCN experiencing a decline of $0.05\%$ and MQ-GraphSAGE a decline of $0.2\%$. The staleness caused by asynchronous updates during gradient sharing and model updates causes this decline. RaCoM does not eliminate the effects of staleness, especially in models that are more susceptible to asynchronous updates. However, it reduces significantly through periodic synchronization and appropriate gradient management.

For smaller datasets like ogbn-proteins, MQ-GraphSAGE provides $2.61\,\times$ improvement. Batch time decreases proportionally, emphasizing MQ-GNN's ability to adapt to varying graph sizes. Because of the sparsity of the graphs and infrequent model synchronization, MQ-GNN exhibits a more considerable decrease in metrics for smaller datasets like ogbn-proteins and ogbn-arxiv. On ogbn-proteins, MQ-GraphSAGE's evaluation metric decreased by $0.3\%$, whereas on ogbn-arxiv, it decreased by $0.28\%$ (Table~\ref{tab:g2-ns-results}). The sparse connectivity of these datasets reduces the benefits of caching and queuing, and infrequent synchronization increases the impact of staleness in asynchronous updates, degrading overall model accuracy.

Scaling up to three and four GPUs, MQ-GNN continues to outperform baselines in training and batch times, as evidenced by Tables~\ref{tab:g3-ns-results} to \ref{tab:g4-ladies-large-results} in Appendix~\ref{app2:sota_mq_gnn_results_3_4}. For the Reddit dataset, MQ-GCN delivers a remarkable \(3.95\,\times\) improvement on three GPUs, as shown in Table~\ref{tab:g3-ns-results}. Similarly, for the ogbn-products dataset, MQ-GraphSAGE achieved a \(3.36\,\times\) speedup. For the ogbn-arxiv dataset, MQ-GraphSAGE achieves a \(2.89\,\times\) speedup in training time. The batch time decreases significantly, from $31.59$ ms to $10.45$ ms (Table~\ref{tab:g3-ns-results}). With four GPUs, MQ-GNN maintained its efficiency, as seen on the ogbn-arxiv dataset, where MQ-GCN achieved a \(3.45\,\times\) improvement in training time (Table~\ref{tab:g4-ns-large-results}). On dense datasets such as Reddit and ogbn-products, MQ-GNN excels because it leverages caching and overlapping for optimal efficiency. MQ-GNN's architecture efficiently lowers synchronization and communication overhead, especially in dense graphs. These findings show that MQ-GNN can efficiently manage dense datasets, even though synchronization becomes more difficult as the number of GPUs increases.

\begin{table*}[] 
\caption{Results for batch time (ms), training time (ms), and evaluation metrics (\%) for FastGCN, MQ-FastGCN, and their enhanced variants on two-GPU configuration.\label{tab:g2-fastgcn-large-results}}
\begin{tabular}{llllll}
\hline
Model                            & Benchmark & ogbn-proteins           & ogbn-arxiv              & Reddit                           & ogbn-products                    \\ \hline
\multirow{3}{*}{FastGCN}     & Batch     & 8.04 $\pm$ 0.37         & 7.46 $\pm$ 0.28         & 9.08 $\pm$ 0.41                  & 10.36 $\pm$ 0.45                 \\
                                 & Training  & 137406.91 $\pm$ 1421.17 & 133517.13 $\pm$ 1402.34 & \textbf{273813.32 $\pm$ 2845.98} & 151858.83 $\pm$ 1559.87          \\
                                 & Metrics   & 53.21 $\pm$ 0.48        & 26.19 $\pm$ 0.39        & 45.81 $\pm$ 0.55                 & 28.88 $\pm$ 0.52                 \\ \hline
\multirow{3}{*}{MQ-FastGCN}      & Batch     & 2.90 $\pm$ 8.21         & 2.61 $\pm$ 7.89         & 2.72 $\pm$ 10.45                 & 3.15 $\pm$ 9.34                  \\
                                 & Training  & 51189.87 $\pm$ 0.55     & 48299.73 $\pm$ 0.40     & \textbf{83994.41 $\pm$ 0.60}     & 47605.60 $\pm$ 0.49              \\
                                 & Metrics   & 52.71 $\pm$ 0.40        & 25.89 $\pm$ 0.35        & 45.71 $\pm$ 0.48                 & 28.58 $\pm$ 0.45                 \\ \hline
\multirow{3}{*}{FastGCN+f}   & Batch     & 8.04 $\pm$ 0.42         & 7.87 $\pm$ 0.39         & 8.93 $\pm$ 0.48                  & 8.78 $\pm$ 0.43                  \\
                                 & Training  & 137441.27 $\pm$ 1457.12 & 140884.09 $\pm$ 1482.94 & 269335.06 $\pm$ 2798.42          & 128694.32 $\pm$ 1351.89          \\
                                 & Metrics   & 61.12 $\pm$ 0.48        & 26.20 $\pm$ 0.65        & 53.64 $\pm$ 0.69                 & 33.24 $\pm$ 0.63                 \\ \hline
\multirow{3}{*}{MQ-FastGCN+f}    & Batch     & 3.04 $\pm$ 8.32         & 2.91 $\pm$ 8.45         & 2.75 $\pm$ 10.75                 & 2.98 $\pm$ 9.15                  \\
                                 & Training  & 52834.47 $\pm$ 0.52     & 53127.50 $\pm$ 0.45     & 84174.25 $\pm$ 0.63              & 44628.97 $\pm$ 0.58              \\
                                 & Metrics   & 60.72 $\pm$ 0.40        & 26.0 $\pm$ 0.50         & 53.54 $\pm$ 0.60                 & 33.04 $\pm$ 0.50                 \\ \hline
\multirow{3}{*}{FastGCN+d}   & Batch     & 8.05 $\pm$ 0.39         & 8.06 $\pm$ 0.35         & 8.79 $\pm$ 0.47                  & 9.59 $\pm$ 0.52                  \\
                                 & Training  & 137557.27 $\pm$ 1435.89 & 144348.22 $\pm$ 1487.32 & 265117.67 $\pm$ 2759.83          & 140487.23 $\pm$ 1453.76          \\
                                 & Metrics   & 53.70 $\pm$ 0.72        & 27.37 $\pm$ 0.68        & 44.28 $\pm$ 0.83                 & 28.01 $\pm$ 1.43                 \\ \hline
\multirow{3}{*}{MQ-FastGCN+d}    & Batch     & 3.12 $\pm$ 8.57         & 3.07 $\pm$ 8.94         & 2.75 $\pm$ 11.10                 & 3.35 $\pm$ 9.78                  \\
                                 & Training  & 54547.94 $\pm$ 0.62     & 55924.65 $\pm$ 0.65     & 84382.63 $\pm$ 0.73              & 50125.19 $\pm$ 0.85              \\
                                 & Metrics   & 53.20 $\pm$ 0.60        & 27.07 $\pm$ 0.60        & 44.18 $\pm$ 0.70                 & 27.71 $\pm$ 1.20                 \\ \hline
\multirow{3}{*}{FastGCN+f+d} & Batch     & 7.67 $\pm$ 0.43         & 7.66 $\pm$ 0.35         & 8.88 $\pm$ 0.50                  & 9.59 $\pm$ 0.34                  \\
                                 & Training  & 131084.62 $\pm$ 1348.91 & 137053.24 $\pm$ 1421.38 & 267829.32 $\pm$ 2823.94          & \textbf{140487.28 $\pm$ 1459.75} \\
                                 & Metrics   & 61.15 $\pm$ 0.12        & 24.32 $\pm$ 0.18        & 53.90 $\pm$ 0.19                 & \textbf{30.14 $\pm$ 0.37}        \\ \hline
\multirow{3}{*}{MQ-FastGCN+f+d}  & Batch     & 2.82 $\pm$ 8.11         & 2.77 $\pm$ 8.45         & 2.69 $\pm$ 10.98                 & 3.22 $\pm$ 9.05                  \\
                                 & Training  & 49439.57 $\pm$ 0.50     & 50670.86 $\pm$ 0.45     & 82618.40 $\pm$ 0.58              & \textbf{48081.08 $\pm$ 0.55}     \\
                                 & Metrics   & 60.75 $\pm$ 0.10        & 24.12 $\pm$ 0.15        & 53.80 $\pm$ 0.12                 & \textbf{29.94 $\pm$ 0.35}        \\ \hline
\end{tabular}
\end{table*}

\begin{table*}[] 
\caption{Results for batch time (ms), training time (ms), and evaluation metrics (\%) for  LADIES, MQ-LADIES, and their enhanced variants on two-GPU configuration.\label{tab:g2-ladies-large-results}}
\begin{tabular}{llllll}
\hline
                                 & Benchmark & ogbn-proteins           & ogbn-arxiv              & Reddit                           & ogbn-products                                     \\ \hline
\multirow{3}{*}{LADIES}      & Batch     & 6.85 $\pm$ 0.19         & 10.92 $\pm$ 0.41        & 8.20 $\pm$ 0.33                  & 7.59 $\pm$ 0.28                                   \\
                                 & Training  & 117160.16 $\pm$ 1198.45 & 235377.73 $\pm$ 2430.52 & \textbf{191979.81 $\pm$ 1998.45} & 366542.86 $\pm$ 3721.24                           \\
                                 & Metrics   & 68.34 $\pm$ 0.16        & 60.62 $\pm$ 0.23        & 75.47 $\pm$ 0.22                 & 53.47 $\pm$ 0.42                                  \\ \hline
\multirow{3}{*}{MQ-LADIES}       & Batch     & 2.49 $\pm$ 5.25         & 3.90 $\pm$ 8.90         & 2.49 $\pm$ 11.25                 & 2.41 $\pm$ 9.95                                   \\
                                 & Training  & 43320.33 $\pm$ 0.08     & 85432.90 $\pm$ 0.11     & \textbf{59420.67 $\pm$ 0.13}     & 118779.52 $\pm$ 0.15                              \\
                                 & Metrics   & 67.94 $\pm$ 0.10        & 60.42 $\pm$ 0.20        & 75.42 $\pm$ 0.18                 & 53.27 $\pm$ 0.40                                  \\ \hline
\multirow{3}{*}{LADIES+f}    & Batch     & 6.72 $\pm$ 0.24         & 10.60 $\pm$ 0.43        & 8.43 $\pm$ 0.48                  & 8.23 $\pm$ 0.36                                   \\
                                 & Training  & 114955.31 $\pm$ 1203.42 & 222753.14 $\pm$ 2299.01 & \textbf{198578.13 $\pm$ 2011.68} & \textbf\{438722.13\} $\pm$ 4487.24 \\
                                 & Metrics   & 68.66 $\pm$ 0.07        & 63.45 $\pm$ 0.14        & \textbf{90.55 $\pm$ 0.16}        & \textbf{63.21 $\pm$ 0.22}                         \\ \hline
\multirow{3}{*}{MQ-LADIES+f}     & Batch     & 2.44 $\pm$ 5.15         & 3.75 $\pm$ 8.65         & 2.53 $\pm$ 10.95                 & 2.53 $\pm$ 10.25                                  \\
                                 & Training  & 42378.54 $\pm$ 0.09     & 80344.90 $\pm$ 0.13     & \textbf{60673.42 $\pm$ 0.15}     & \textbf{137522.97 $\pm$ 0.22}                     \\
                                 & Metrics   & 68.36 $\pm$ 0.08        & 63.25 $\pm$ 0.15        & \textbf{90.50 $\pm$ 0.12}        & \textbf{63.01 $\pm$ 0.20}                         \\ \hline
\multirow{3}{*}{LADIES+d}    & Batch     & 11.05 $\pm$ 0.53        & 13.68 $\pm$ 0.39        & 9.84 $\pm$ 0.47                  & 7.85 $\pm$ 0.34                                   \\
                                 & Training  & 188865.71 $\pm$ 1924.32 & 327562.53 $\pm$ 3321.40 & 443242.73 $\pm$ 4520.24          & 451953.42 $\pm$ 4615.22                           \\
                                 & Metrics   & 68.77 $\pm$ 0.11        & 61.43 $\pm$ 0.17        & 87.71 $\pm$ 0.18                 & 55.58 $\pm$ 0.37                                  \\ \hline
\multirow{3}{*}{MQ-LADIES+d}     & Batch     & 4.03 $\pm$ 8.35         & 5.01 $\pm$ 10.78        & 3.08 $\pm$ 12.45                 & 2.53 $\pm$ 11.35                                  \\
                                 & Training  & 70227.20 $\pm$ 0.12     & 122060.70 $\pm$ 0.14    & 141574.51 $\pm$ 0.18             & 147443.91 $\pm$ 0.23                              \\
                                 & Metrics   & 68.37 $\pm$ 0.12        & 61.13 $\pm$ 0.18        & 87.61 $\pm$ 0.18                 & 55.28 $\pm$ 0.35                                  \\ \hline
\multirow{3}{*}{LADIES+f+d} & Batch     & 11.78 $\pm$ 0.41        & 12.80 $\pm$ 0.34        & 10.69 $\pm$ 0.50                 & 7.80 $\pm$ 0.42                                   \\
                                 & Training  & 201339.29 $\pm$ 2034.67 & 280377.04 $\pm$ 2863.15 & 481194.71 $\pm$ 4895.74          & 452177.30 $\pm$ 4610.84                           \\
                                 & Metrics   & 67.56 $\pm$ 0.16        & 62.99 $\pm$ 0.18        & 88.94 $\pm$ 0.24                 & 63.53 $\pm$ 0.48                                  \\ \hline
\multirow{3}{*}{LADIES+f+d}  & Batch     & 4.52 $\pm$ 8.95         & 4.77 $\pm$ 10.95        & 3.23 $\pm$ 13.10                 & 2.67 $\pm$ 12.15                                  \\
                                 & Training  & 78953.76 $\pm$ 0.11     & 106808.64 $\pm$ 0.15    & 147969.36 $\pm$ 0.20             & 158074.05 $\pm$ 0.28                              \\
                                 & Metrics   & 67.16 $\pm$ 0.15        & 62.79 $\pm$ 0.18        & 88.84 $\pm$ 0.20                 & 63.33 $\pm$ 0.45                                  \\ \hline
\end{tabular}
\end{table*}

\textbf{LS Performance}
MQ-GNN achieved significant speed improvements over layer-wise approaches for two GPUs, as shown in Tables~\ref{tab:g2-fastgcn-large-results} and~\ref{tab:g2-ladies-large-results}. On the Reddit dataset using MQ-FastGCN, the training time improves by \(3.26\,\times\) (Table~\ref{tab:g2-fastgcn-large-results}). Similarly, MQ-LADIES provides a \(3.23\,\times \) reduction in training time (Table~\ref{tab:g2-ladies-large-results}). On the ogbn-products dataset, MQ-FastGCN+f+d achieved a \(2.92\,\times\) improvement, with batch times decreasing by \(2.97\,\times\) and a slight $0.2\%$ drop in the metric. For dense graphs like Reddit, MQ-LADIES+f experiences a smaller decrease in its evaluation metric, with a $0.05\%$ drop compared to its counterpart, while achieving a \(3.27\,\times\) speedup. These results emphasize MQ-GNN's ability to integrate additional computational overhead, such as flat sampling  \eqref{eq:calibProFlat} and debiasing \eqref{eq:calibProDeb}, in advanced variants of LADIES and FastGCN, with minimal runtime overhead.

As the number of GPUs increases to three and four, MQ-GNN continues to show remarkable efficiency, as shown in Tables~\ref{tab:g3-ns-results}-\ref{tab:g4-ladies-large-results} in Appendix~\ref{app2:sota_mq_gnn_results_3_4}. On three GPUs, MQ-FastGCN achieves a \(3.38\,\times \) improvement, as presented in Table~\ref{tab:g3-fastgcn-large-results}. Similarly, on the ogbn-arxiv dataset, MQ-LADIES reduces training time by \(2.95\,\times\). Batch time follows a similar trend, dropping from $10.92$ ms to $3.52$ ms (Table~\ref{tab:g3-ladies-results}). Despite the training speedup, evaluation metrics experience a slightly larger drop due to staleness, with MQ-FastGCN+f+d dropping the metric by $0.8\%$ (Table~\ref{tab:g3-fastgcn-large-results}). MQ-LADIES+f+d demonstrates strong scalability on four GPUs, achieving a \(3.6\,\times\) reduction in training time for the ogbn-products dataset. Batch time improves proportionally, dropping from $8.36$ ms to $2.23$ ms at the cost of a $1.0\%$ reduction in the metric, showcasing MQ-GNN's ability to dynamically balance staleness and synchronization costs, as presented in Table~\ref{tab:g4-ladies-large-results}. On the Reddit dataset, MQ-LADIES+f+d and MQ-FastGCN+f+d provide a \(4.35\,\times\) speedup in training, with a slight $0.6\%$ drop in the metric (Tables ~\ref{tab:g4-ladies-large-results} and~\ref{tab:g4-fastgcn-large-results}). Even with three and four GPUs, MQ-GNN maintains evaluation metrics close to its baselines, with only minor deviations due to reduced staleness through periodic synchronization using RaCoM (Section~\ref{sec:model_update_gpus}). Furthermore, the integration of additional computational overhead, such as flat sampling \eqref{eq:calibProFlat} and debiasing \eqref{eq:calibProDeb} in LADIES+f+d and FastGCN+f+d, is efficiently managed within MQ-GNN's pipeline without significantly increasing runtime. This is achieved by efficiently overlapping these tasks with other operations. Certain models, such as LADIES, are more sensitive to staleness caused by asynchronous updates, which affect evaluation metrics, particularly in scenarios with infrequent synchronization.

\textbf{Comparative Analysis} By aligning updates at regular intervals, RaCoM ensures that the model remains synchronized, even in scenarios with high communication demands. This approach allows MQ-GNN to deliver stable evaluation metrics across configurations with two or three GPUs, with slight declines when scaling to four GPUs, particularly on smaller or sparse datasets, such as ogbn-proteins.

Graph sparsity amplifies the effects of staleness because the periodic synchronization intervals in RaCoM cannot fully address the asynchronous updates required for consistent feature aggregation. For instance, in \eqref{eq:gcn_eq}, staleness can disrupt the aggregation process when it uses outdated node embeddings ($Z^{l-1}$) due to delayed gradient updates. This leads to inconsistent feature propagation, which is especially problematic in sparse datasets, where every edge and feature has a more pronounced impact on the overall aggregation. Similarly, in GraphSAGE, staleness disrupts the sampling process when neighbor embeddings \eqref{eq:sage_eq} are not synchronized. Sparse graphs worsen this issue because of a limited number of neighbors, making embedding of each neighbor critical. Thus, RaCoM's synchronization periods do not align well with updates, resulting in inconsistent aggregation and degraded performance.

Furthermore, staleness in the RaCoM model also affects LS approaches. For instance, FastGCN, as defined in \eqref{eq:fastgcn}, samples nodes at each layer, assuming independence among sampled nodes. Thus, delayed synchronization impacts FastGCN because it struggles with outdated sampled embeddings that fail to represent the current state of the graph, leading to estimation errors and high variance in the aggregation process. Dense datasets partially mitigate this issue due to redundancy, but sparse graphs amplify the problem as fewer nodes are sampled. Compared to FastGCN, LADIES is more severely affected because each layer depends on sampled nodes from the previous layer, as defined in \eqref{eq:ladies}. The hierarchical dependency between layers further magnifies the effects of staleness. When embedding updates from earlier layers are asynchronous, the approximation quality in subsequent layers suffers, particularly in sparse graphs. MQ-GNN's periodic synchronization through RaCoM does not align effectively with the layer-dependent sampling of LADIES, resulting in higher estimation errors and degraded performance. While flat sampling \eqref{eq:calibProFlat} and debiasing \eqref{eq:calibProDeb} techniques for FastGCN and LADIES aim to improve sampling robustness by reducing variance and correcting estimation biases, their effectiveness diminishes in multi-GPU environments where delays in gradient synchronization introduce non-uniformity in sampling distributions. This introduces another layer of complexity that is difficult for MQ-GNN's current synchronization strategy to address completely.

In multi-GPU systems, the pipelining method with multi-queuing in MQ-GNN is essential for maximizing resource efficiency. The method enqueues a predetermined number of mini-batches to reduce GPU idle time and guarantee seamless task transitions. The queuing strategy improves training efficiency and performs well across various datasets and configurations. This method works well for large datasets, where scalability depends on sustaining high throughput across several GPUs.

Integrating these techniques—RaCoM, multi-queue, and GNS—makes MQ-GNN a robust framework for multi-GPU setups. The periodic synchronization provided by RaCoM mitigates the impact of asynchronous processing and communication delays. The multi-queue approach enhances GPU utilization by decreasing idle time and guaranteeing seamless training. By reducing communication overhead in data transfer to GPU, GNS further accelerates the training. Together, these techniques allow MQ-GNN to achieve scalable and efficient performance, albeit with a slight drop in metrics across large and small datasets. These improvements show MQ-GNN's effectiveness as a scalable alternative to existing baselines in multi-GPU environments. While MQ-GNN provides significant computational speedups and scalability, its synchronization and queuing mechanisms are less effective at stabilizing the performance of layer-wise sampling models under multi-GPU training.

\section{Conclusion}
This paper presented MQ-GNN, a scalable and efficient framework for multi-GPU GNN training. MQ-GNN employs multi-queue pipelining to efficiently overlap mini-batch generation, data transfer, and computation, reducing idle times and maximizing resource utilization. It incorporates an adaptive queue-sizing strategy to balance memory constraints and computational efficiency while leveraging global neighbor sampling with caching to minimize data transfer overhead. Furthermore, the RaCoM periodic synchronization minimizes communication bottlenecks while maintaining model accuracy considering the graph characteristics and number of GPUs. 

Experimental results on four large-scale datasets and ten baseline models demonstrate that MQ-GNN delivers up to \(4.6\,\times\) faster training and \( 30\%\) higher GPU utilization, with only a \( 2\%\) accuracy trade-off. These results validate MQ-GNN's effectiveness in optimizing large-scale GNN training across diverse graph structures.

MQ-GNN offers a high-performance and scalable graph learning solution for applications that include recommendation engines, social networks, and biological systems. Future research will extend MQ-GNN to dynamic graphs and more complex GNN structures to further improve its suitability for changing large-scale datasets.

\appendices

\section{\break Periodic Synchronization}\label{app:periodic_sych}
Training GNNs involves three primary factors: node-wise computations, edge-wise computations, and GPU workload. Node-wise computations scale with \( |V| \), contributing to the workload through node embeddings and local feature transformations. Larger \( |V| \) requires a longer synchronization interval to balance computational cost while maintaining efficiency. Edge-wise computations scale with \( |E| \), contributing to neighbor aggregation and communication overhead. The square root of \( |E| \) models the diminishing returns of additional edges on staleness, ensuring that denser graphs synchronize more frequently to prevent divergence caused by asynchronous updates. GPU workload distribution further influences synchronization; as \( \left| \mathcal{G} \right| \) increases, the workload is spread across more devices, shortening the synchronization interval to enhance consistency across distributed systems.

\begin{proposition}\label{prop:periodic_interval}
 The optimal periodic synchronization interval \( P_{ms} \), which balances computational efficiency and model consistency, is given by:

\begin{equation}
P_{ms} = \left\lceil \frac{\sqrt{|V|}}{\sqrt{\left| \mathcal{G} \right| |E|}} \right\rceil
\end{equation}

This interval ensures:
\begin{itemize}
    \item  The synchronization frequency dynamically adapts to the graph density.
   \item  The trade-off between staleness (due to asynchronous updates) and synchronization overhead (caused by frequent communication) is minimized.
\end{itemize}

\end{proposition}

\begin{proof}
Let \( C(P) \) be the total cost function that balances staleness and synchronization costs:

\[
C(P) = \alpha P  |E| + \beta  \frac{1}{P}  \frac{|V|}{\left| \mathcal{G} \right|},
\]

where \( P|E| \) represents the staleness cost, which increases linearly with \( P \) and is proportional to the number of edges \( |E| \). The term \( \frac{1}{P} \frac{|V|}{\left| \mathcal{G} \right|} \) represents the synchronization cost, which decreases inversely with \( P \) and is proportional to the number of nodes \( |V| \), divided by the number of GPUs \( \left| \mathcal{G} \right| \). The constants \(\alpha\) and \(\beta\) are weighting coefficients that control the relative contribution of each term.

To minimize \( C(P) \), we differentiate \( C(P) \) with respect to \( P \) and set \( \frac{dC(P)}{dP} = 0 \):

\[
\frac{dC(P)}{dP} = \alpha|E| - \beta \frac{1}{P^2} \frac{|V|}{\left| \mathcal{G} \right|}
\]

Rearranging terms gives:

\[
\alpha |E| = \beta  \frac{1}{P^2} \frac{|V|}{\left| \mathcal{G} \right|}
\]

Multiplying through by \( P^2 \) yields:

\[
P^2 = \frac{\beta}{\alpha} \frac{|V|}{\left| \mathcal{G} \right||E|}
\]

Taking the square root of both sides gives:

\[
P = \sqrt{\frac{\beta}{\alpha}}\sqrt{\frac{|V|}{\left| \mathcal{G} \right||E|}}
\]

Here, \( \sqrt{\frac{\beta}{\alpha}} \) is treated as a constant scaling factor \( k \). For simplicity, normalize \( k = 1 \), resulting in:

\[
P = \frac{\sqrt{|V|}}{\sqrt{\left| \mathcal{G} \right||E|}}
\]

To ensure \( P \) is a discrete value suitable for implementation, apply the ceiling function:

\[
P_{ms} = \left\lceil \frac{\sqrt{|V|}}{\sqrt{\left| \mathcal{G} \right||E|}} \right\rceil
\]

This derivation shows that \( P_{ms} \) dynamically adapts to the structure of the graph \( G \) (\( |V| \) and \( |E| \)) and the number of GPUs (\(  \left| \mathcal{G} \right| \)), ensuring a balance between staleness and synchronization costs.

\end{proof}

\section{\centering Results Comparison with SoTA for Three, and Four GPUs} \label{app2:sota_mq_gnn_results_3_4}

Tables~\ref{tab:g3-ns-results} to~\ref{tab:g4-ladies-large-results} present the comparison results for three and four GPU configurations, demonstrating MQ-GNN’s scalability and efficiency. The results highlight training speedup, batch time improvements, and the impact of staleness on evaluation metrics, showing how MQ-GNN balances performance and accuracy through periodic synchronization.

\begin{table*}[t] 
\caption{Results for batch time (ms), training time (ms), and evaluation metrics (\%) for GCN, GraphSAGE, MQ-GCN, and MQ-GraphSAGE on a three-GPU configuration.\label{tab:g3-ns-results}}
\begin{tabular}{llllll}
\hline
                               & Benchmark & ogbn-proteins           & ogbn-arxiv                      & Reddit                           & ogbn-products                   \\ \hline
\multirow{3}{*}{GCN}       & Batch     & 13.58 $\pm$ 0.12        & 13.85 $\pm$ 0.10                & 24.25 $\pm$ 0.18                 & 15.19 $\pm$ 0.11                \\
                               & Training  & 155635.53 $\pm$ 1525.12 & 249346.67 $\pm$ 2050.21         & \textbf{506732.17 $\pm$ 3810.48} & 296226.39 $\pm$ 1820.34         \\
                               & Metrics   & 66.79 $\pm$ 0.70        & 68.45 $\pm$ 0.50                & 90.82 $\pm$ 0.35                 & 71.04 $\pm$ 0.42                \\ \hline
\multirow{3}{*}{MQ-GCN}        & Batch     & 4.52 $\pm$ 0.18         & 4.39 $\pm$ 0.25                 & 5.91 $\pm$ 0.30                  & 4.22 $\pm$ 0.20                 \\
                               & Training  & 54487.44 $\pm$ 2400.35  & \textbf{83083.30 $\pm$ 3260.28} & \textbf{128234.18 $\pm$ 5210.45} & \textbf{85834.73 $\pm$ 2100.48} \\
                               & Metrics   & 66.29 $\pm$ 0.35        & 68.15 $\pm$ 0.42                & 90.62 $\pm$ 0.60                 & 70.74 $\pm$ 0.50                \\ \hline
\multirow{3}{*}{GraphSAGE} & Batch     & 25.01 $\pm$ 0.64        & \textbf{31.59 $\pm$ 0.43}       & 38.29 $\pm$ 0.26                 & 23.65 $\pm$ 0.59                \\
                               & Training  & 277099.73 $\pm$ 0.16    & \textbf{432195.54 $\pm$ 0.28}   & 777316.19 $\pm$ 0.94             & \textbf{378204.24 $\pm$ 0.05}   \\
                               & Metrics   & 66.62 $\pm$ 0.98        & 69.07 $\pm$ 0.47                & 94.53 $\pm$ 0.95                 & 63.72$\pm$ 0.26                 \\ \hline
\multirow{3}{*}{MQ-GraphSAGE}  & Batch     & 8.50 $\pm$ 4.12         & \textbf{10.45 $\pm$ 5.85}       & 9.45 $\pm$ 7.15                  & 6.76 $\pm$ 5.05                 \\
                               & Training  & 98610.54 $\pm$ 110.25   & \textbf{149490.10 $\pm$ 150.35} & 199308.66 $\pm$ 220.18           & \textbf{112520.80 $\pm$ 175.45} \\
                               & Metrics   & 65.92 $\pm$ 0.32        & 67.66 $\pm$ 0.38                & 90.83 $\pm$ 0.15                 & 70.74 $\pm$ 0.45                \\ \hline
\end{tabular}
\end{table*}

\begin{table*}[t] 
\caption{Results for batch time (ms), training time (ms), and evaluation metrics (\%) for  FastGCN, MQ-FastGCN, and their enhanced variants on three-GPU configuration.\label{tab:g3-fastgcn-large-results}}
\begin{tabular}{llllll}
\hline
                                 & Benchmark & ogbn-proteins                  & ogbn-arxiv            & Reddit                & ogbn-products                  \\ \hline
\multirow{3}{*}{FastGCN}     & Batch     & 7.82 $\pm$ 0.35                & 8.09 $\pm$ 0.30       & 9.21 $\pm$ 0.40       & 9.62 $\pm$ 0.35                \\
                                 & Training  & 89664.64 $\pm$ 725             & 97615.04 $\pm$ 810    & 185078.34 $\pm$ 1290  & \textbf{94393.48 $\pm$ 520}    \\
                                 & Metrics   & 53.69 $\pm$ 0.35               & 26.9 $\pm$ 0.42       & 43.61 $\pm$ 0.50      & 28.81 $\pm$ 0.45               \\ \hline
\multirow{3}{*}{MQ-FastGCN}      & Batch     & 2.74 $\pm$ 1.50                & 2.70 $\pm$ 1.80       & 2.30 $\pm$ 3.25       & 2.73 $\pm$ 2.40                \\
                                 & Training  & 33072.93 $\pm$ 125.35          & 34224.56 $\pm$ 135.45 & 48058.41 $\pm$ 155.25 & \textbf{27917.82 $\pm$ 145.40} \\
                                 & Metrics   & 52.69 $\pm$ 0.20               & 26.1 $\pm$ 0.25       & 43.31 $\pm$ 0.25      & 28.11 $\pm$ 0.40               \\ \hline
\multirow{3}{*}{FastGCN+f}   & Batch     & 7.69 $\pm$ 0.20                & 7.81 $\pm$ 0.15       & 8.54 $\pm$ 0.25       & 9.02 $\pm$ 0.18                \\
                                 & Training  & 88148.54 $\pm$ 680             & 70788.87 $\pm$ 590    & 173517.37 $\pm$ 1200  & 89131.48 $\pm$ 610             \\
                                 & Metrics   & 62.1 $\pm$ 0.40                & 22.46 $\pm$ 0.35      & 52.99 $\pm$ 0.40      & 32.29 $\pm$ 0.32               \\ \hline
\multirow{3}{*}{MQ-FastGCN+f}    & Batch     & 2.66 $\pm$ 1.70                & 2.56 $\pm$ 2.00       & 2.08 $\pm$ 3.18       & 2.52 $\pm$ 2.20                \\
                                 & Training  & 32371.82 $\pm$ 130.25          & 24400.47 $\pm$ 110.25 & 43924.31 $\pm$ 145.30 & 26060.72 $\pm$ 125.45          \\
                                 & Metrics   & 61.3 $\pm$ 0.22                & 21.86 $\pm$ 0.25      & 52.69 $\pm$ 0.22      & 31.69 $\pm$ 0.42               \\ \hline
\multirow{3}{*}{FastGCN+d}   & Batch     & 7.62 $\pm$ 0.06                & 7.70 $\pm$ 0.37       & 8.75 $\pm$ 0.87       & 9.54 $\pm$ 1.05                \\
                                 & Training  & 87327.23 $\pm$ 0.96            & 69783.84 $\pm$ 0.62   & 175894.47 $\pm$ 0.92  & 93658.32 $\pm$ 0.05            \\
                                 & Metrics   & 54.22 $\pm$ 1.04               & 27.0 $\pm$ 0.76       & 44.93 $\pm$ 0.29      & 30.09 $\pm$ 0.38               \\ \hline
\multirow{3}{*}{MQ-FastGCN+d}    & Batch     & 2.71 $\pm$ 1.75                & 2.64 $\pm$ 1.90       & 2.24 $\pm$ 3.45       & 2.80 $\pm$ 2.75                \\
                                 & Training  & 32623.50 $\pm$ 125.45          & 25098.34 $\pm$ 130.25 & 46716.02 $\pm$ 145.80 & 28718.33 $\pm$ 135.45          \\
                                 & Metrics   & 53.32 $\pm$ 0.25               & 26.30 $\pm$ 0.30      & 44.53 $\pm$ 0.20      & 29.29 $\pm$ 0.40               \\ \hline
\multirow{3}{*}{FastGCN+f+d} & Batch     & 7.85 $\pm$ 0.22                & 8.01 $\pm$ 0.74       & 8.91$\pm$ 0.7         & 9.54$\pm$ 0.95                 \\
                                 & Training  & \textbf{89942.67 $\pm$ 1.01}   & 96594.43 $\pm$ 0.01   & 179902.72 $\pm$ 0.38  & 93632.14 $\pm$ 0.64            \\
                                 & Metrics   & \textbf{61.76 $\pm$ 0.07}      & 24.25 $\pm$ 0.2       & 54.53 $\pm$ 0.96      & 32.23 $\pm$ 0.88               \\ \hline
\multirow{3}{*}{MQ-FastGCN+f+d}  & Batch     & 2.77 $\pm$ 1.80                & 2.67 $\pm$ 2.15       & 2.20 $\pm$ 3.30       & 2.72 $\pm$ 2.50                \\
                                 & Training  & \textbf{33275.04 $\pm$ 135.40} & 33642.75 $\pm$ 140.20 & 45964.57 $\pm$ 150.30 & 27860.07 $\pm$ 130.35          \\
                                 & Metrics   & \textbf{60.96 $\pm$ 0.15}      & 23.75 $\pm$ 0.20      & 54.23 $\pm$ 0.20      & 31.73 $\pm$ 0.35               \\ \hline
\end{tabular}
\end{table*}

\begin{table*}[t] 
\caption{Results for batch time (ms), training time (ms), and evaluation metrics (\%) for  LADIES, MQ-LADIES, and their enhanced variants on three-GPU configuration.\label{tab:g3-ladies-results}}
\begin{tabular}{llllll}
\hline
                                & Benchmark & ogbn-proteins           & ogbn-arxiv                       & Reddit                  & ogbn-products           \\ \hline
\multirow{3}{*}{LADIES}     & Batch     & 6.85 $\pm$ 0.20         & 10.92 $\pm$ 0.35                 & 8.20 $\pm$ 0.30         & 7.59 $\pm$ 0.25         \\
                                & Training  & 117160.16 $\pm$ 910.25  & \textbf{235377.73 $\pm$ 1850.52} & 191979.81 $\pm$ 1510.34 & 366542.86 $\pm$ 2760.28 \\
                                & Metrics   & 68.34 $\pm$ 0.18        & 60.62 $\pm$ 0.28                 & 75.47 $\pm$ 0.16        & 53.47 $\pm$ 0.30        \\ \hline
\multirow{3}{*}{MQ-LADIES}      & Batch     & 2.38 $\pm$ 0.54         & 3.52 $\pm$ 0.58                  & 1.97 $\pm$ 0.16         & 2.08 $\pm$ 0.47         \\
                                & Training  & 42568.80 $\pm$ 980.25   & \textbf{79730.09 $\pm$ 1785.50}  & 47994.18 $\pm$ 1687.45  & 104707.20 $\pm$ 1520.37 \\
                                & Metrics   & 67.44 $\pm$ 0.15        & 60.02 $\pm$ 0.33                 & 75.17 $\pm$ 0.37        & 52.97 $\pm$ 0.74        \\ \hline
\multirow{3}{*}{LADIES+f}   & Batch     & 6.02 $\pm$ 0.20         & 10.30 $\pm$ 0.30                 & 8.51 $\pm$ 0.35         & 8.12 $\pm$ 0.15         \\
                                & Training  & 68952.23 $\pm$ 520.10   & 134775.56 $\pm$ 950.45           & 209233.72 $\pm$ 1320.75 & 150464.29 $\pm$ 920.30  \\
                                & Metrics   & 68.67 $\pm$ 0.12        & 63.30 $\pm$ 0.20                 & 90.87 $\pm$ 0.25        & 62.41 $\pm$ 0.18        \\ \hline
\multirow{3}{*}{MQ-LADIES+f}    & Batch     & 2.06 $\pm$ 0.42         & 3.26 $\pm$ 0.38                  & 2.02 $\pm$ 0.18         & 2.19 $\pm$ 0.22         \\
                                & Training  & 24675.93 $\pm$ 920.45   & 44787.82 $\pm$ 1187.35           & 51639.00 $\pm$ 1520.62  & 42384.17 $\pm$ 1387.44  \\
                                & Metrics   & 67.87 $\pm$ 0.13        & 62.80 $\pm$ 0.23                 & 90.67 $\pm$ 0.15        & 62.01 $\pm$ 0.30        \\ \hline
\multirow{3}{*}{LADIES+d}   & Batch     & 10.87 $\pm$ 0.35        & 12.34 $\pm$ 0.40                 & 9.97 $\pm$ 0.28         & 8.02 $\pm$ 0.20         \\
                                & Training  & 124582.76 $\pm$ 980.15  & 222182.65 $\pm$ 1450.52          & 286160.53 $\pm$ 1980.85 & 137120.38 $\pm$ 1150.30 \\
                                & Metrics   & 69.30 $\pm$ 0.15        & 62.29 $\pm$ 0.18                 & 88.43 $\pm$ 0.14        & 54.91 $\pm$ 0.24        \\ \hline
\multirow{3}{*}{MQ-LADIES+d}    & Batch     & 3.80 $\pm$ 0.38         & 4.18 $\pm$ 0.28                  & 2.49 $\pm$ 0.45         & 2.30 $\pm$ 0.22         \\
                                & Training  & 45733.19 $\pm$ 1250.45  & 78674.80 $\pm$ 1487.34           & 74325.03 $\pm$ 2287.64  & 41063.36 $\pm$ 1350.52  \\
                                & Metrics   & 68.40 $\pm$ 0.15        & 61.69 $\pm$ 0.28                 & 88.13 $\pm$ 0.26        & 54.31 $\pm$ 0.42        \\ \hline
\multirow{3}{*}{LADIES+f+d} & Batch     & 11.82 $\pm$ 0.40        & 12.88 $\pm$ 0.35                 & 9.97 $\pm$ 0.28         & 8.01 $\pm$ 0.20         \\
                                & Training  & 135476.01 $\pm$ 1080.34 & 231789.48 $\pm$ 1630.25          & 229440.32 $\pm$ 1970.56 & 131150.32 $\pm$ 1140.78 \\
                                & Metrics   & 68.06 $\pm$ 0.12        & 60.01 $\pm$ 0.18                 & 88.78 $\pm$ 0.15        & 62.11 $\pm$ 0.22        \\ \hline
\multirow{3}{*}{MQ-LADIES+f+d}  & Batch     & 4.22 $\pm$ 0.28         & 4.43 $\pm$ 0.38                  & 2.52 $\pm$ 0.12         & 2.35 $\pm$ 0.28         \\
                                & Training  & 50542.12 $\pm$ 1025.35  & 83561.08 $\pm$ 1820.44           & 60232.61 $\pm$ 2350.54  & 40353.69 $\pm$ 1425.87  \\
                                & Metrics   & 67.26 $\pm$ 0.23        & 59.51 $\pm$ 0.72                 & 88.48 $\pm$ 0.25        & 61.61 $\pm$ 0.35        \\ \hline
\end{tabular}
\end{table*}

\begin{table*}[t] 
\caption{Results for batch time (ms), training time (ms), and evaluation metrics (\%) for GCN, GraphSAGE, MQ-GCN, and MQ-GraphSAGE on a four-GPU configuration.\label{tab:g4-ns-large-results}}
\begin{tabular}{llllll}
\hline
                               & Benchmark & ogbn-proteins           & ogbn-arxiv                       & Reddit                           & ogbn-products           \\ \hline
\multirow{3}{*}{GCN}       & Batch     & 13.17 $\pm$ 0.15        & 14.48 $\pm$ 0.09                 & 22.42 $\pm$ 0.85                 & 15.70 $\pm$ 0.38        \\
                               & Training  & 113834.39 $\pm$ 845.32  & \textbf{173942.63 $\pm$ 1378.41} & 351391.48 $\pm$ 2825.67          & 226895.51 $\pm$ 1545.27 \\
                               & Metrics   & 66.81 $\pm$ 0.80        & 68.54 $\pm$ 0.40                 & 90.97 $\pm$ 0.70                 & 71.40 $\pm$ 0.45        \\ \hline
\multirow{3}{*}{MQ-GCN}        & Batch     & 3.87 $\pm$ 0.25         & 4.02 $\pm$ 0.30                  & 4.87 $\pm$ 0.50                  & 3.92 $\pm$ 0.45         \\
                               & Training  & 34954.27 $\pm$ 265.15   & \textbf{50400.06 $\pm$ 290.27}   & 78948.03 $\pm$ 400.67            & 58925.16 $\pm$ 335.28   \\
                               & Metrics   & 65.61 $\pm$ 0.40        & 67.74 $\pm$ 0.45                 & 90.47 $\pm$ 0.55                 & 70.80 $\pm$ 0.50        \\ \hline
\multirow{3}{*}{GraphSAGE} & Batch     & 20.54 $\pm$ 0.32        & 32.08 $\pm$ 0.25                 & 39.36 $\pm$ 0.60                 & 24.38 $\pm$ 0.50        \\
                               & Training  & 177562.71 $\pm$ 1260.32 & 353692.29 $\pm$ 2058.45          & \textbf{602149.33 $\pm$ 3450.87} & 298715.49 $\pm$ 1885.40 \\
                               & Metrics   & 66.62 $\pm$ 0.65        & 69.33 $\pm$ 0.50                 & \textbf{94.79 $\pm$ 0.30}        & 63.85 $\pm$ 0.40        \\ \hline
\multirow{3}{*}{MQ-GraphSAGE}  & Batch     & 6.14 $\pm$ 0.50         & 9.14 $\pm$ 0.60                  & 8.65 $\pm$ 0.70                  & 6.25 $\pm$ 0.60         \\
                               & Training  & 55442.75 $\pm$ 385.45   & 105458.24 $\pm$ 500.62           & \textbf{136716.51 $\pm$ 670.54}  & 79655.60 $\pm$ 450.27   \\
                               & Metrics   & 65.12 $\pm$ 0.45        & 68.33 $\pm$ 0.50                 & \textbf{94.19 $\pm$ 0.35}        & 62.95 $\pm$ 0.42        \\ \hline
\end{tabular}
\end{table*}

\begin{table*}[t] 
\caption{Results for batch time (ms), training time (ms), and evaluation metrics (\%) for FastGCN, MQ-FastGCN, and their enhanced variants on four-GPU configuration.\label{tab:g4-fastgcn-large-results}}
\begin{tabular}{llllll}
\hline
                                 & Benchmark & ogbn-proteins         & ogbn-arxiv            & Reddit                          & ogbn-products          \\ \hline
\multirow{3}{*}{FastGCN}     & Batch     & 7.85 $\pm$ 0.97       & 7.84 $\pm$ 0.85       & 9.24 $\pm$ 0.75                 & 12.75 $\pm$ 0.58       \\
                                 & Training  & 67839.25 $\pm$ 525.45 & 53289.46 $\pm$ 395.25 & 139282.58 $\pm$ 865.35          & 125574.71 $\pm$ 710.15 \\
                                 & Metrics   & 53.69 $\pm$ 0.50      & 26.90 $\pm$ 0.55      & 43.67 $\pm$ 0.35                & 27.01 $\pm$ 0.25       \\ \hline
\multirow{3}{*}{MQ-FastGCN}      & Batch     & 2.41 $\pm$ 0.25       & 2.30 $\pm$ 0.30       & 2.05 $\pm$ 0.38                 & 3.35 $\pm$ 0.42        \\
                                 & Training  & 21868.93 $\pm$ 165.25 & 16377.15 $\pm$ 150.35 & 31989.47 $\pm$ 195.50           & 34383.46 $\pm$ 215.40  \\
                                 & Metrics   & 51.69 $\pm$ 0.32      & 25.40 $\pm$ 0.38      & 42.97 $\pm$ 0.30                & 25.81 $\pm$ 0.48       \\ \hline
\multirow{3}{*}{FastGCN+f}   & Batch     & 7.57 $\pm$ 0.45       & 7.90 $\pm$ 0.55       & 8.61 $\pm$ 0.52                 & 9.36 $\pm$ 0.65        \\
                                 & Training  & 65406.74 $\pm$ 495.30 & 53655.51 $\pm$ 425.10 & 129790.42 $\pm$ 875.30          & 92650.35 $\pm$ 640.45  \\
                                 & Metrics   & 62.10 $\pm$ 0.65      & 24.46 $\pm$ 0.45      & 53.12 $\pm$ 0.50                & 32.32 $\pm$ 0.40       \\ \hline
\multirow{3}{*}{MQ-FastGCN+f}    & Batch     & 2.30 $\pm$ 0.28       & 2.29 $\pm$ 0.32       & 1.87 $\pm$ 0.45                 & 2.43 $\pm$ 0.55        \\
                                 & Training  & 20741.16 $\pm$ 170.20 & 16211.68 $\pm$ 155.50 & 29163.13 $\pm$ 205.75           & 25026.60 $\pm$ 185.40  \\
                                 & Metrics   & 60.40 $\pm$ 0.32      & 23.16 $\pm$ 0.35      & 52.52 $\pm$ 0.45                & 31.22 $\pm$ 0.50       \\ \hline
\multirow{3}{*}{FastGCN+d}   & Batch     & 7.45 $\pm$ 0.45       & 7.66 $\pm$ 0.40       & 8.49 $\pm$ 0.58                 & 9.38 $\pm$ 0.55        \\
                                 & Training  & 64367.37 $\pm$ 470.50 & 52058.51 $\pm$ 400.50 & 128041.19 $\pm$ 860.25          & 92877.09 $\pm$ 645.35  \\
                                 & Metrics   & 61.78 $\pm$ 0.60      & 24.14 $\pm$ 0.55      & 54.74 $\pm$ 0.60                & 32.27 $\pm$ 0.45       \\ \hline
\multirow{3}{*}{MQ-FastGCN+d}    & Batch     & 2.33 $\pm$ 0.28       & 2.32 $\pm$ 0.32       & 1.93 $\pm$ 0.42                 & 2.53 $\pm$ 0.50        \\
                                 & Training  & 20934.90 $\pm$ 180.20 & 16419.90 $\pm$ 165.50 & 30123.73 $\pm$ 215.00           & 26154.32 $\pm$ 150.00  \\
                                 & Metrics   & 59.98 $\pm$ 0.35      & 22.74 $\pm$ 0.40      & 54.04 $\pm$ 0.50                & 30.97 $\pm$ 0.55       \\ \hline
\multirow{3}{*}{FastGCN+f+d} & Batch     & 7.45 $\pm$ 0.40       & 7.66 $\pm$ 0.35       & 8.49 $\pm$ 0.52                 & 9.38 $\pm$ 0.50        \\
                                 & Training  & 64367.37 $\pm$ 455.35 & 52058.51 $\pm$ 385.25 & \textbf{128041.19 $\pm$ 825.50} & 92877.09 $\pm$ 620.15  \\
                                 & Metrics   & 61.78 $\pm$ 0.55      & 24.14 $\pm$ 0.50      & \textbf{54.74 $\pm$ 0.50}       & 32.27 $\pm$ 0.40       \\ \hline
\multirow{3}{*}{MQ-FastGCN+f+d}  & Batch     & 2.31 $\pm$ 0.25       & 2.29 $\pm$ 0.28       & 1.88 $\pm$ 0.35                 & 2.50 $\pm$ 0.42        \\
                                 & Training  & 20793.48 $\pm$ 165.25 & 16162.30 $\pm$ 150.35 & \textbf{29399.34 $\pm$ 210.50}  & 25782.21 $\pm$ 185.25  \\
                                 & Metrics   & 60.18 $\pm$ 0.30      & 22.94 $\pm$ 0.35      & \textbf{54.14 $\pm$ 0.40}       & 31.27 $\pm$ 0.50       \\ \hline
\end{tabular}
\end{table*}

\begin{table*}[t] 
\caption{Results for batch time (ms), training time (ms), and evaluation metrics (\%) for LADIES, MQ-LADIES, and their enhanced variants on four-GPU configuration.\label{tab:g4-ladies-large-results}}
\begin{tabular}{llllll}
\hline
                                & Benchmark & ogbn-proteins          & ogbn-arxiv              & Reddit                           & ogbn-products                    \\ \hline
\multirow{3}{*}{LADIES}     & Batch     & 6.46 $\pm$ 0.28        & 10.18 $\pm$ 0.38        & 8.85 $\pm$ 0.45                  & 8.37 $\pm$ 0.42                  \\
                                & Training  & 55798.51 $\pm$ 485.20  & 126832.24 $\pm$ 1025.75 & 139342.54 $\pm$ 905.80           & 123057.34 $\pm$ 925.35           \\
                                & Metrics   & 69.30 $\pm$ 0.55       & 62.34 $\pm$ 0.65        & 75.70 $\pm$ 0.50                 & 54.00 $\pm$ 0.60                 \\ \hline
\multirow{3}{*}{MQ-LADIES}      & Batch     & 1.97 $\pm$ 1.45        & 2.91 $\pm$ 1.20         & 1.88 $\pm$ 0.50                  & 2.11 $\pm$ 0.65                  \\
                                & Training  & 17756.09 $\pm$ 110.25  & 37821.10 $\pm$ 165.45   & 30600.62 $\pm$ 125.80            & 32376.30 $\pm$ 165.60            \\
                                & Metrics   & 67.50 $\pm$ 0.25       & 61.04 $\pm$ 0.65        & 75.10 $\pm$ 0.37                 & 53.00 $\pm$ 0.95                 \\ \hline
\multirow{3}{*}{LADIES+f}   & Batch     & 6.81 $\pm$ 0.30        & 10.44 $\pm$ 0.40        & 11.00 $\pm$ 0.50                 & 8.18 $\pm$ 0.35                  \\
                                & Training  & 58849.62 $\pm$ 520.35  & 102393.24 $\pm$ 850.40  & 173280.67 $\pm$ 1020.25          & 120272.43 $\pm$ 900.20           \\
                                & Metrics   & 68.47 $\pm$ 0.45       & 64.10 $\pm$ 0.55        & 91.08 $\pm$ 0.50                 & 62.31 $\pm$ 0.55                 \\ \hline
\multirow{3}{*}{MQ-LADIES+f}    & Batch     & 2.05 $\pm$ 1.25        & 2.94 $\pm$ 1.00         & 2.31 $\pm$ 0.75                  & 2.04 $\pm$ 0.50                  \\
                                & Training  & 18479.93 $\pm$ 125.72  & 30093.14 $\pm$ 160.65   & 37667.94 $\pm$ 145.50            & 31208.25 $\pm$ 180.40            \\
                                & Metrics   & 66.97 $\pm$ 0.95       & 63.00 $\pm$ 0.45        & 90.58 $\pm$ 0.50                 & 61.51 $\pm$ 0.35                 \\ \hline
\multirow{3}{*}{LADIES+d}   & Batch     & 11.90 $\pm$ 0.50       & 12.11 $\pm$ 0.40        & 9.47 $\pm$ 0.60                  & 8.31 $\pm$ 0.45                  \\
                                & Training  & 102820.34 $\pm$ 810.35 & 163524.31 $\pm$ 1325.40 & 213050.34 $\pm$ 1650.50          & 122178.21 $\pm$ 1050.35          \\
                                & Metrics   & 69.29 $\pm$ 0.55       & 62.30 $\pm$ 0.45        & 88.44 $\pm$ 0.50                 & 54.89 $\pm$ 0.60                 \\ \hline
\multirow{3}{*}{MQ-LADIES+d}    & Batch     & 3.66 $\pm$ 1.00        & 3.56 $\pm$ 0.90         & 2.06 $\pm$ 0.50                  & 2.16 $\pm$ 0.30                  \\
                                & Training  & 33037.28 $\pm$ 150.50  & 50276.67 $\pm$ 190.85   & 47833.83 $\pm$ 175.65            & 32997.83 $\pm$ 180.90            \\
                                & Metrics   & 67.59 $\pm$ 1.00       & 60.90 $\pm$ 0.85        & 87.74 $\pm$ 0.50                 & 53.79 $\pm$ 0.45                 \\ \hline
\multirow{3}{*}{LADIES+f+d} & Batch     & 11.70 $\pm$ 0.50       & 13.03 $\pm$ 0.65        & 11.00 $\pm$ 0.55                 & \textbf{8.36 $\pm$ 0.45}         \\
                                & Training  & 101075.36 $\pm$ 810.45 & 129564.43 $\pm$ 1250.35 & \textbf{247618.18 $\pm$ 1620.65} & \textbf{122857.42 $\pm$ 1035.25} \\
                                & Metrics   & 68.07 $\pm$ 0.45       & 60.00 $\pm$ 0.55        & \textbf{89.43 $\pm$ 0.55}        & 62.60 $\pm$ 0.45                 \\ \hline
\multirow{3}{*}{MQ-LADIES+f+d}  & Batch     & 3.63 $\pm$ 1.20        & 3.88 $\pm$ 1.35         & 2.44 $\pm$ 0.40                  & \textbf{2.23 $\pm$ 0.85}         \\
                                & Training  & 32675.67 $\pm$ 145.85  & 40188.03 $\pm$ 175.90   & \textbf{56834.54 $\pm$ 150.55}   & \textbf{34087.17 $\pm$ 125.50}   \\
                                & Metrics   & 66.47 $\pm$ 0.25       & 58.80 $\pm$ 0.85        & \textbf{88.83 $\pm$ 0.60}        & 61.60 $\pm$ 0.20                 \\ \hline
\end{tabular}
\end{table*}

\clearpage

\section*{Acknowledgment}
This work was supported by the Institute of Information \& communications Technology Planning \& Evaluation (IITP) grant funded by the Korean government (MSIT) (No.2021-0-00859, Development of a distributed graph DBMS for intelligent processing of big graphs.

This work was supported by the Institute of Information \& communications Technology Planning \& Evaluation (IITP) grant funded by the Korean government(MSIT) (No.RS-2022-00155911, Artificial Intelligence Convergence Innovation Human Resources Development (Kyung Hee University)).

\bibliographystyle{IEEEtran}
\bibliography{mq_gnn}

\begin{IEEEbiography}[{\includegraphics[width=1in,height=1.25in,clip,keepaspectratio]{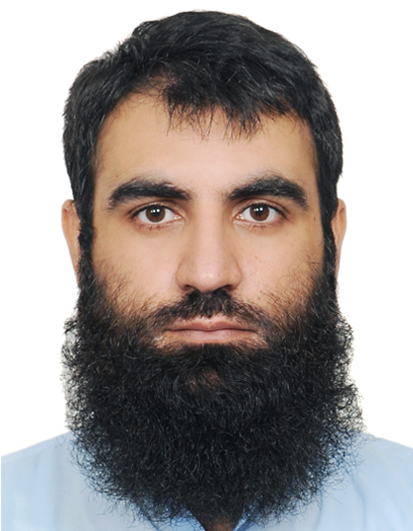}}]{Irfan Ullah} received a Master’s degree in Computer Science from the National University of Sciences and Technology (NUST), Pakistan. He is currently pursuing a Ph.D. degree in Computer Science at Kyung Hee University, South Korea, where he is also working as a Research Assistant in the Department of Computer Science and Engineering.

His research interests include graph neural networks, machine learning, deep learning, big data analytics, distributed computing, social computing, natural language processing, operating system design, and memory system optimization. He has contributed to multiple research projects on scalable GNN training, memory-aware computing, distributed graph learning, and social media data analysis. His work has been published in high-impact journals and international conferences, and he has authored several patents related to NLP and GNN.

Previously, he held research and teaching positions at NUST, Federal Urdu University of Arts, Science, and Technology (FUUAST), and the multi-national company Knowledge Platform, Pakistan. He has also been actively involved in open-source projects. He has delivered talks and tutorials on programming and development, natural language processing, distributed computing, machine learning, and machine learning with graphs.

\end{IEEEbiography}

\begin{IEEEbiography}[{\includegraphics[width=1in,height=1.25in,clip,keepaspectratio]{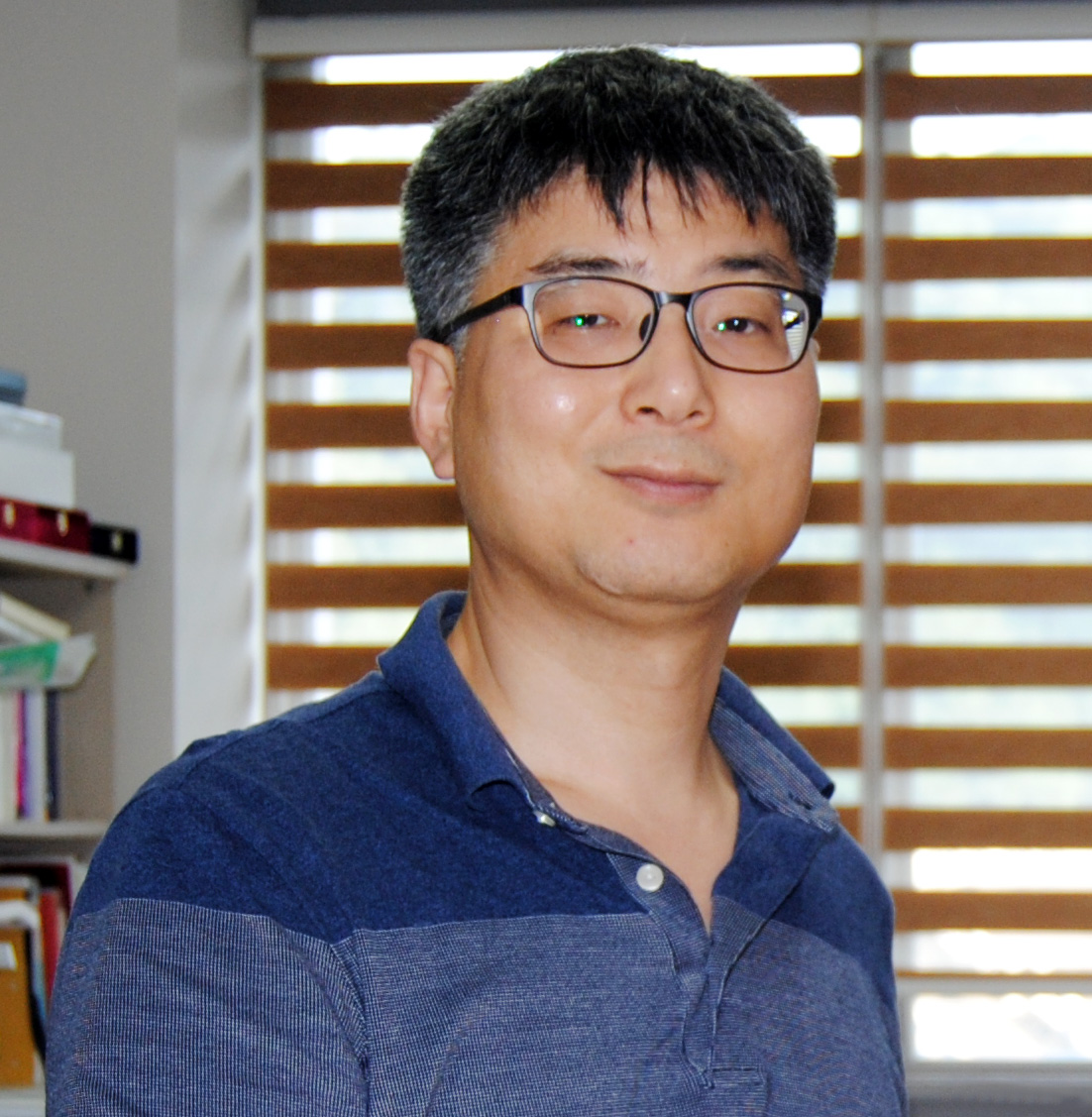}}]{Young-Koo Lee} (Member, IEEE) received the B.S., M.S., and Ph.D. degrees in Computer Science from the Korea Advanced Institute of Science and Technology (KAIST), Daejeon, South Korea, in 1992, 1994, and 2002, respectively.  

From 2002 to 2004, he was a Postdoctoral Researcher at the University of Illinois at Urbana-Champaign and a Postdoctoral Fellow at the Advanced Information Technology Research Center (AITrc), KAIST. In 2004, he joined Kyung Hee University, Global Campus, South Korea, where he has held various academic positions. He served as an Assistant Professor from 2004 to 2010, an Associate Professor from 2010 to 2015, and has been a Professor since 2015 at the College of Software. His research focuses on graph neural networks (GNNs), data mining, online analytical processing, query optimization, and big data processing.

Prof. Lee has published numerous research papers in top-tier journals and conferences. He has actively contributed to advancing data science and GNN-based learning methodologies. He is a member of IEEE and has received recognition for his contributions to large-scale graph processing and deep-learning applications.
\end{IEEEbiography}

\EOD

\end{document}